\documentclass{article}
\usepackage{rotating}
\usepackage[final]{neurips_2022}
\usepackage{graphicx}
\usepackage{booktabs} % To thicken table lines
\usepackage{array}
\usepackage{wrapfig}
\usepackage{multirow}
\usepackage{float}
\usepackage{rotating}
\usepackage{tikz}
\usepackage{stfloats}
\usepackage{amsmath}
\usepackage{natbib}
\setcitestyle{numbers,sort&compress, super}
\usepackage{makecell}
\usepackage{xcolor,colortbl}
\usepackage{nicematrix}
\usepackage{enumitem}
\usepackage{amsmath}
\usepackage{titlesec}
\DeclareMathOperator*{\argmax}{arg\,max}
\newcommand{\CC}[1]{\cellcolor{gainsboro!#1}}
\newcommand{\revfig}[1]{#1}
\newcommand{\revtext}[1]{\textcolor{black}{#1}}

\newcommand{\newmet}[0]{MED}

\newcommand{\citenst}[1]{\citeauthor{#1}}
\usepackage[utf8]{inputenc} % allow utf-8 input
\usepackage[T1]{fontenc}    % use 8-bit T1 fonts
\usepackage{hyperref}       % hyperlinks
\usepackage{url}            % simple URL typesetting
\usepackage{booktabs}       % professional-quality tables
\usepackage{amsfonts}       % blackboard math symbols
\usepackage{nicefrac}       % compact symbols for 1/2, etc.
\usepackage{microtype}      % microtypography
\usepackage{xcolor}         % colors
\usepackage{subcaption}
\usepackage[font=small,labelfont=bf]{caption}
\titlespacing{\section}{0pt}{0.5\parskip}{-0.3\parskip}
\titlespacing{\subsection}{0pt}{0.5\parskip}{-0.3\parskip}
\titlespacing{\subsubsection}{0pt}{0.5\parskip}{-0.3\parskip}

\begin{document}

\title{An Empirical Study on Disentanglement of Negative-free Contrastive Learning}

\author{%
  Jinkun Cao$^3$ \hspace{0.5cm} Ruiqian Nai$^1$\hspace{0.5cm} Qing Yang$^4$\hspace{0.5cm} Jialei Huang$^1$\hspace{0.5cm}  Yang Gao$^{1,2}$\thanks{corresponding author}\\
  Tsinghua University$^1$ \hspace{0.2cm}  Shanghai Qi-Zhi Institute$^2$ \hspace{0.2cm} \\Carnegie Mellon University$^3$ \hspace{0.2cm} Shanghai Jiao Tong University$^4$\\
}

\setlength{\abovedisplayskip}{1pt}
\setlength{\belowdisplayskip}{1pt}

\maketitle
\vspace{-1em}
\begin{abstract}
Negative-free contrastive learning methods have attracted a lot of attention with simplicity and impressive performances for large-scale pretraining. However, its disentanglement property remains unexplored. In this paper, we examine negative-free contrastive learning methods to study the disentanglement property empirically. We find that existing disentanglement metrics fail to make meaningful measurements for high-dimensional representation models, so we propose a new disentanglement metric based on Mutual Information between latent representations and data factors. With this proposed metric, we benchmark the disentanglement property of negative-free contrastive learning on both popular synthetic datasets and a real-world dataset CelebA. Our study shows that the investigated methods can learn a well-disentangled subset of representation. As far as we know, we are the first to extend the study of disentangled representation learning to high-dimensional representation space and introduce negative-free contrastive learning methods into this area. The source code of this paper is available at \url{https://github.com/noahcao/disentanglement_lib_med}.
\end{abstract}

\section{Introduction}
Existing disentangled representation learning methods are mostly generative models~\cite{kingma2013auto,higgins2016beta,kim2018disentangling,goodfellow2014generative,chen2016infogan,lin2020infogan}. They are evaluated only on simple synthetic datasets~\cite{dsprites17,3dshapes18} and low representation dimensions, e.g., no more than 20-d.
In contrast, contrastive learning is a class of discriminative methods, trained by pulling the representation of two augmentations of the same image close. This usually requires a much higher representation dimension, e.g., often at least 1000-d.

Despite the success of contrastive learning, it remains unknown whether it can lead to disentangled representations.
Recently, some works~\cite{wang2020understanding,zimmermann2021contrastive} reveal that contrastive learning can approximately invert the data generation process and allow the learned representation have identifiability, which is related to the disentanglement property. 
However, all these advances necessarily rely on the contrast provided by negative samples. The disentanglement property of contrastive learning without negatives, or ``non-contrastive self-supervised learning'', remains unexplored. \revtext{And no contrastive learning methods have been evaluated on the standard disentanglement benchmarks~\cite{dsprites17,3dshapes18} yet.}

Representation disentanglement requires different dimensions of a representation vector to be correlated to only one independent variable of the input data, which we call a data factor. There lacks a uniform convention of measuring of ``correlation'' here, so multiple disentanglement metrics~\cite{eastwood2018framework,higgins2016beta,halva2021disentangling} have been proposed for the quantitative evaluation. Their shared principle is that a well-disentangled representation model should not have a representation dimension responding to the change of more than one data factor.
We find although multiple disentanglement metrics'~\cite{higgins2016beta,kim2018disentangling,chen2018isolating,eastwood2018framework,kumar2017variational} evaluations are in agreement for low-dimensional space~\cite{locatello2019challenging}, they disagree in high-dimensional space. Moreover, their design limitation makes the measurement not meaingful in high-dimensional space.
We thus propose a new disentanglement evaluation metric that is robust to representation dimension scaling and is thus adaptable to a high-dimensional representation model. The new metric is named as ``\textbf{M}utual information based \textbf{E}ntroy \textbf{D}isentanglement'' and MED for short. 

With the proposed metric, we find that negative-free contrastive learning methods can achieve good disentanglement in a subset of latent dimensions. Given that the recent disentanglement studies are limited to simple synthetic datasets but contrastive learning is especially powerful for complicated and large-scale datasets, we also extend the quantitative benchmarking to a real-world dataset CelebA~\cite{liu2018large}. On CelebA, existing low-dimensional generative disentangled learning methods are unable to learn good representations, demonstrating the gap between current disentangled learning research and the real-world data complexity. To summarize, our contributions in the work are three-fold:

1. We find that existing disentanglement metrics fail to extend to high-dimensional representation space, and we propose a new metric MED to extend this area to high-dimensional space.

2. We extend the study of disentangled representation learning to real-world complicated datasets and high-dimensional representation space, revealing the gap between current disentangled representation learning and real-world data complexity.

3. We empirically study the disentanglement property of contrastive learning without negatives. We find it can learn a well-disentangled subspace of latent representation.

\section{Related Works}
\textbf{Disentangled Representation Learning.}  
Learning a disentangled representation is a long-desired goal in the deep learning community~\cite{bengio2013representation, peters2017elements, goodfellow2016deep, bengio2007scaling, schmidhuber1992learning,lake2017building, tschannen2018recent}. 
A disentangled representation matches how humans understand the world and allows us to use much fewer labels to learn challenging downstream tasks~\cite{van2019disentangled}. 
There are two lines of work related to this goal, i.e., the Independent Component Analysis (ICA) and the Disentangled Deep Representation Learning.
ICA~\citep{ICAbook} usually assumes that the pattern of noise~\citep{hyvarinen2016unsupervised, khemakhem2020variational} or some additional auxiliary variables~\citep{hyvarinen2019nonlinear, icebeem} can be observed. 
On the other hand, deep representation learning makes no explicit assumption on the noise distribution or representation prior and usually emphasizes unsupervised learning. Under this setting, deep representation learning is usually based on deep generative models such as VAE-based methods~\citep{higgins2016beta, kim2018disentangling, chen2018isolating, kumar2017variational} and Generative Adversarial Networks (GAN)~\citep{goodfellow2014generative, chen2016infogan}. 
These two lines of study focus on different but relevant~\citep{slowvae} aspects of a learned representation. In this paper, we focus on empirically studying the disentanglement property of negative-free contrastive learning methods to introduce them into the scope of disentangled representation learning.

\textbf{Disentanglement Metrics.}  
The variation in the metric used also shows the difference between ICA and Disentangled Deep Representation Learning.
ICA aims to achieve good identifiability and uses Mean Correlation Coefficient (MCC) as the common metric. 
On the other hand, the metrics used in the deep disentangled representation learning community are very diverse. 
Until now, no widely accepted definition of ``disentanglement'' is available. So the empirical agreement of metrics makes the basis of quantitatively evaluating disentanglement.
DisLib~\citep{locatello2019challenging} summarizes six popular disentanglement metrics, i.e., DCI~\citep{eastwood2018framework}, SAP~\citep{kumar2017variational}, MIG~\citep{chen2018isolating}, BetaVAE score~\citep{higgins2016beta}, FactorVAE score~\cite{kim2018disentangling} and Modularity~\citep{modularity}. 
DisLib finds that the five metrics except for Modularity have good agreement in evaluating disentanglement quality.
However, all of the metrics are evaluated on low-dimensional latent space, which is around 10 dimensions. 
We find severe problems when applying those metrics to high-dimensional latent spaces where they show significant disagreement. 
To make meaningful evaluation of the disentanglement property of contrastive learning methods, we thus propose a new metric, which is more applicable to high-dimensional representation space.

\textbf{Contrastive Learning and Representation Disentanglement.}  
Contrastive learning (CL) creates ``views'' by augmentations over images. Views of the same image serve as positives, and views of other images as negatives. 
Recently, some works try to understand contrastive learning theoretically~\citep{wang2020understanding, arora2019theoretical,tsai2020demystifying, tosh2021contrastive, lee2020predicting} or empirically~\citep{tian2020makes,zhao2020makes,purushwalkam2020demystifying}. \citenst{zimmermann2021contrastive} suggests that the contrastive method inverts a data generation process when infinite negative samples are available~\citep{wang2020understanding}, which is related to the disentanglement property. 
\revtext{However, there is still no work connecting CL methods with standard disentanglement benchmarks.}  
On the other hand, negative-free CL methods~\cite{richemond2020byol,simsiam,zbontar2021barlow} do not use the contrast between positive and negative samples to encourage discriminative representations. They generate positive ``views'' of data by applying different augmentations to the same input data, and the pair of views is forwarded into two network streams. 
Recent works on this line have their own unique designs. 
BYOL~\cite{richemond2020byol} discovers that self-supervised learning can avoid trivial solutions, i.e., ``model collapse'', even without using negative samples to provide contrast. The key of BYOL is to add a predictor layer following the commonly adopted ``encoder-projector'' network of contrastive learning methods~\cite{he2020momentum}. This provides additional asymmetry. 
As a follow-up, SimSiam~\cite{simsiam} removes the momentum update from BYOL and proves that ``predictor+stop-gradient'' is enough for self-supervised learning to learn non-trivial representation. 
More recently, Barlow Twins~\cite{zbontar2021barlow} shows that even the predictor or the trick of stop-gradient is not necessary. 
Negative-free CL methods have their own advantages, such as avoiding model collapse~\cite{tian2020understanding,hua2021feature,jing2021understanding}, but its disentanglement property remains unexplored, either empirically or theoretically. In this paper, our focus thus to benchmark the disentanglement property of these methods for the first time which is possible after we propose a new disentanglement metric applicable to the high-dimensional representation models.

\section{The Proposed Disentanglement Metric: MED}
Positive-negative contrast in self-supervised learning is seen to encourage learned representation to have uniformity on a hyper-sphere~\cite{wang2020understanding}. When negative views are not available, the disentanglement property of the learned representation remains a mystery.
Therefore, we examine the mentioned negative-free contrastive learning methods in an empirical study to reveal this property of interest.
However, we demonstrate in Section~\ref{sec:failure_existing} that existing disentanglement metrics can not evaluate CL methods in high-dimensional space fairly nor meaningfully. Therefore, we introduce our proposed MI-based Entropy Disentanglement score (\newmet) in Section~\ref{sec:med} and its variant to evaluate disentanglement of a subspace of representation in Section~\ref{sec:partial_med}.

\subsection{Failure of Existing Metrics on High-Dimensional Space}
\label{sec:failure_existing}

Typical contrastive learning methods need a high dimensional representation space (``latent space'') to train well. 

However, the previous study of disentangled representation learning only deals with low-dimensional representation space. For example, in \citenst{locatello2019challenging}, the latent space dimension is set to $10$. So, existing disentanglement metrics (see Appendix~\ref{sec:appendix_more_benchmark} for details) are designed for low-dimensional representation model and have intrinsic flaws when evaluating models in high-dimensional space. To be precise, we have observations as below:

\begin{itemize}[leftmargin=\dimexpr\parindent+0mm+0.5\labelwidth\relax]
    \item Metrics based on learnable classifiers, such as \textbf{BetaVAE score} and \textbf{FactorVAE score}, allow unfair advantages to high-dimensional model whose redundant parameters can trick the classifier more easily. For example, a randomly initialized 1000-d model could reach a FactorVAE score of 61.4 on dSprites, close to many well-trained 10-d VAE-based models' scores (see Table \ref{table:full_results} in Appendix).
    \item Metrics taking only one or two dimensions into score calculation, such as \textbf{SAP} and \textbf{MIG}, are biased to representations of different dimensions. Because a higher dimension makes it harder for an informative dimension to stand out and enjoy a large informativeness gap over other dimensions. 
    \item \textbf{DCI Disentanglement score} uses a learnable regressor to score the importance of each latent dimension to each data factor. The learnable regressor, such as Gradient Boosting Tree (GBT), encourages sparsity in the output importance matrix (see Figure \ref{fig:importance_distri} in Appendix). So, it also gives an advantage to high-dimensional models, making it unfair to compare models of different latent dimensions. Moreover, the construction of regressors is time-intensive in high-dimension space. For example, it usually takes hours to evaluate a 1000-d representation model by DCI using GBT. 
\end{itemize}
These flaws are demonstrated by our experiments, showing that existing metrics disagree for high-dimensional representations. Besides our conceptual justification of the failure of these existing disentanglement metrics, we also construct some scenarios where their failure is theoretically demonstrated in Appendix~\ref{sec:superority_of_med}.
Overall, through our experimental evidence and theoretical justification, existing disentanglement metrics can no longer make meaningful disentanglement measurements in the high-dimensional representation space.

\subsection{\textbf{M}utual Information based \textbf{E}ntropy \textbf{D}isentanglement}
\label{sec:med}
Given the bias and limitations of existing disentanglement metrics and the necessity of high-dimensional representations for contrastive learning, we propose a new disentanglement metric for high dimensional latent spaces, which we name as ``\textbf{M}utual Information based \textbf{E}ntropy \textbf{D}isentanglement", or MED in short. The calculation of MED is based on mutual information (MI) between latent dimensions and the set of data factors of input samples. \revtext{MI is a widely accepted tool to measure correlation of variables and is not biased to models of different dimensions.} Given a dataset generated with $K$ ground truth factors $\boldsymbol{v}\in\mathbb{R}^K$ and a representation vector $\boldsymbol{c} \in \mathbb{R}^D$ , we construct an importance matrix $R \in \mathbb{R}^{D\times K}$ defined by
\begin{equation}
    R_{i j} = I(\boldsymbol{c}_i, \boldsymbol{v}_j) / \sum_{d=0}^{D-1} I(\boldsymbol{c}_d, \boldsymbol{v}_j),
    \label{eq:importance_mat}
\end{equation}
where $I(\boldsymbol{c}_i, \boldsymbol{v}_j)$ denotes the mutual information between the $i^{th}$ latent dimension $\boldsymbol{c}_i$ and the $j^{th}$ ground truth factor $\boldsymbol{v}_j$. Here, each row denotes a representation dimension and each column represents a ground truth factor. We normalize the mutual information by columns, such that an entry in the matrix indicates the relative importance of one dimension over all dimensions regarding a certain data factor. This normalization is necessary  since different dimensions may have different \revtext{overall informativeness to all factors}. \revtext{This operation fixes the gap between models of different dimensions by estimating the relative importance of a single dimension.}

After normalizing over the columns, we evaluate the contribution of a dimension to different factors, which is described by a row of $R$.  If one dimension is informative to only one ground truth factor, then this dimension is \revtext{perfectly} disentangled. 
This matches the mechanism of entropy. So we use the entropy to describe the disentanglement level of a dimension. 
We treat each row as a discrete distribution over factor index by normalization: $P_{i j}=R_{i j} / \sum_{k=0}^{K-1} R_{i k}$, where higher probability $P_{i j}$ indicates that the dimension $\boldsymbol{c}_i$ encodes more information of the factor $\boldsymbol{v}_j$. 
Then the disentanglement score $S_i$ for a latent dimension $\boldsymbol{c}_i$ is calculated as
\begin{equation}
    S_{i}=1-H_{K}\left(P_{i \cdot}\right),
    \label{eq:dis_score}
\end{equation}
where $H_{K}\left(P_{i \cdot}\right) =-\sum_{k=0}^{K-1} P_{i k} \log  P_{i k}$ is the entropy. $S_i$ will be higher if $\boldsymbol{c}_i$ exhibits more informativeness to one factor while less relevance to other factors. Finally, to summarize the overall disentanglement of a representation model, \newmet{} score is the weighted average of the disentanglement scores for all dimensions as
\vspace{-1mm}
\begin{equation}
    \text{MED}\left(\boldsymbol{c}\right) = \sum_{i=0}^{D-1} \rho_{i} S_{i}
\end{equation}
\vspace{-1mm}
where $\rho_i = \sum_{j} R_{i j} / \sum_{i j} R_{i j}$ is the relative importance of each dimension.% summarized on all factors. 

Our proposed MED does not use a learnable classifier nor a regressor. Further, it inherits a DCI-style normalized importance matrix by taking all dimensions into account instead of using only one or two dimensions. These characteristics allow MED to be more robust to the latent dimensionality and to be better suited meaningful disentanglement measurement in high-dimensional space. Besides the advantages MED has, it is also very computationally efficient: for a 1000-d BYOL representation model on Cars3D, the evaluation with DCI with Gradient Boost Tree by DisLib takes more than 14 hours, while MED only takes less than 20 seconds on the same machine.

\subsection{Partial Disentanglement Evaluation Metric}
\label{sec:partial_med}
It is challenging to learn a fully disentangled representation by high-dimensional models without explicitly encouraging disentanglement, especially when there are fewer independent data factors than the number of latent dimensions. However, if a high-dimensional model has a subset of dimensions that disentangle well, it is still worth studying. Such a subset can serve as a proxy to build a more compact representation model when it is hard to train such a model directly.
This motivates us to design a version of \newmet{} to evaluate the partial disentanglement, which we name it as ``Top-k \newmet{}". 

The main difference is that we pick the most disentangled $k$ dimensions for each factor and compute \newmet{} on this subset of representations. 
For each ground truth factor $\boldsymbol{v}_j$,  $\mathcal{G}_j = \left\{i| \argmax_{m}R_{i m} = j\right\}$ is the set of latent dimensions emphasizing this factor. 
Then we pick the top $k$ latent dimensions with the highest disentanglement scores in each $\mathcal{G}_j$ to construct a subset 
$\mathcal{P}_j = \left\{i | S_i \ge S^k, \:i \in \mathcal{G}_j\right\} $. Here, $S_i$ is the disentanglement score of the latent dimension $\boldsymbol{c}_i$ defined in Equation \ref{eq:dis_score}. $S^k$  is the $k^{th}$ highest disentanglement score in $\mathcal{G}_j$. 
Finally  we obtain the subset $\mathcal{P} = \cup _j \mathcal{P}_j$.
We take the sub-vector after selection $\tilde{\boldsymbol{c}} = \left\{\boldsymbol{c}_i\right\}_{i\in\mathcal{P}}$	as the representation to evaluate the \newmet score. The top-k \newmet{} is defined as $\text{MED}(\tilde{\boldsymbol{c}})$.  

\begin{figure}

\begin{subfigure}{\textwidth}
  \centering
  \includegraphics[width=\linewidth,height=75px]{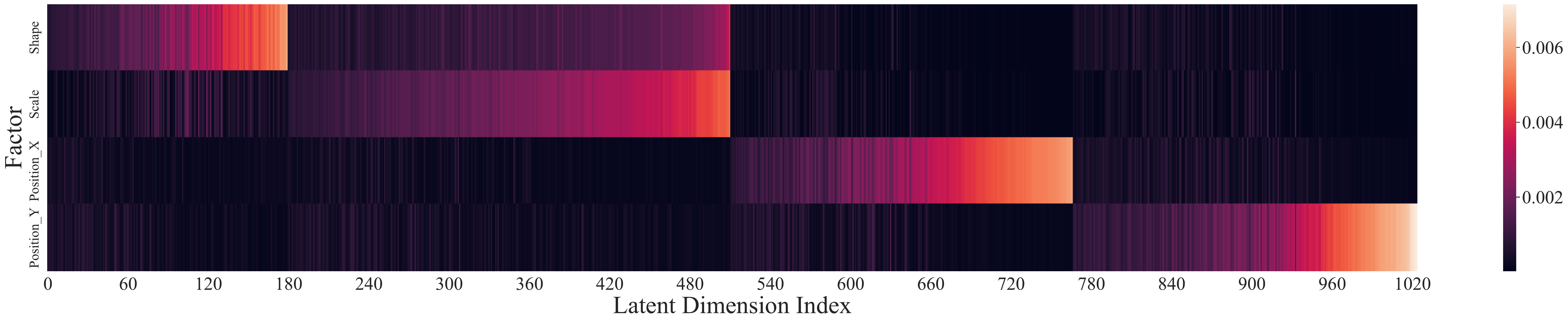}
  \caption{}
  \label{fig:separation}
\end{subfigure}
\\
\begin{subfigure}{\textwidth}
  \centering
  \includegraphics[width=\linewidth,height=75px]{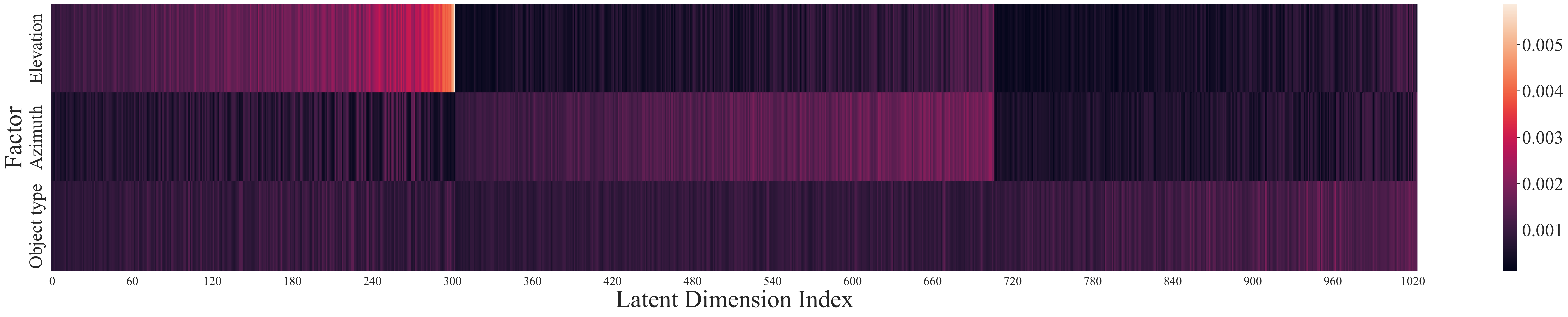}
  \caption{}
  \label{fig:heat_map_cars3d}
\end{subfigure}
\caption{The mutual information heatmap between factors and the BYOL's latent dimensions on dSprites (a) and Cars3D (b). Each row is a ground truth factor and each column is a learned latent dimension. We normalize each row as we do in \newmet{}. Brighter straps denote higher MI. Latent dimensions are sorted for visualization purpose.}
\label{fig:heatmaps}
\vspace{-2em}
\end{figure}

\section{Understanding the learned representation}
\label{sec:understand_exp}
In this part, we qualitatively study the disentanglement of the learned representation by negative-free contrastive learning with BYOL as an example. Since contrastive methods are not generative models, it is hard to directly do factor-controlled pixel-wise reconstruction for visualization. Instead, we measure the mutual information between the learned representations and ground truth factors, which is also the foundation of our proposed MED metric. \revtext{More qualitative study is available in Appendix~\ref{sec:more_qualitative}.}

\subsection{Correspondence of Representation and Factor}
\begin{wrapfigure}{r}{0.3\textwidth}
\vspace{-4em}
  \begin{center}
    \includegraphics[width=0.3\textwidth]{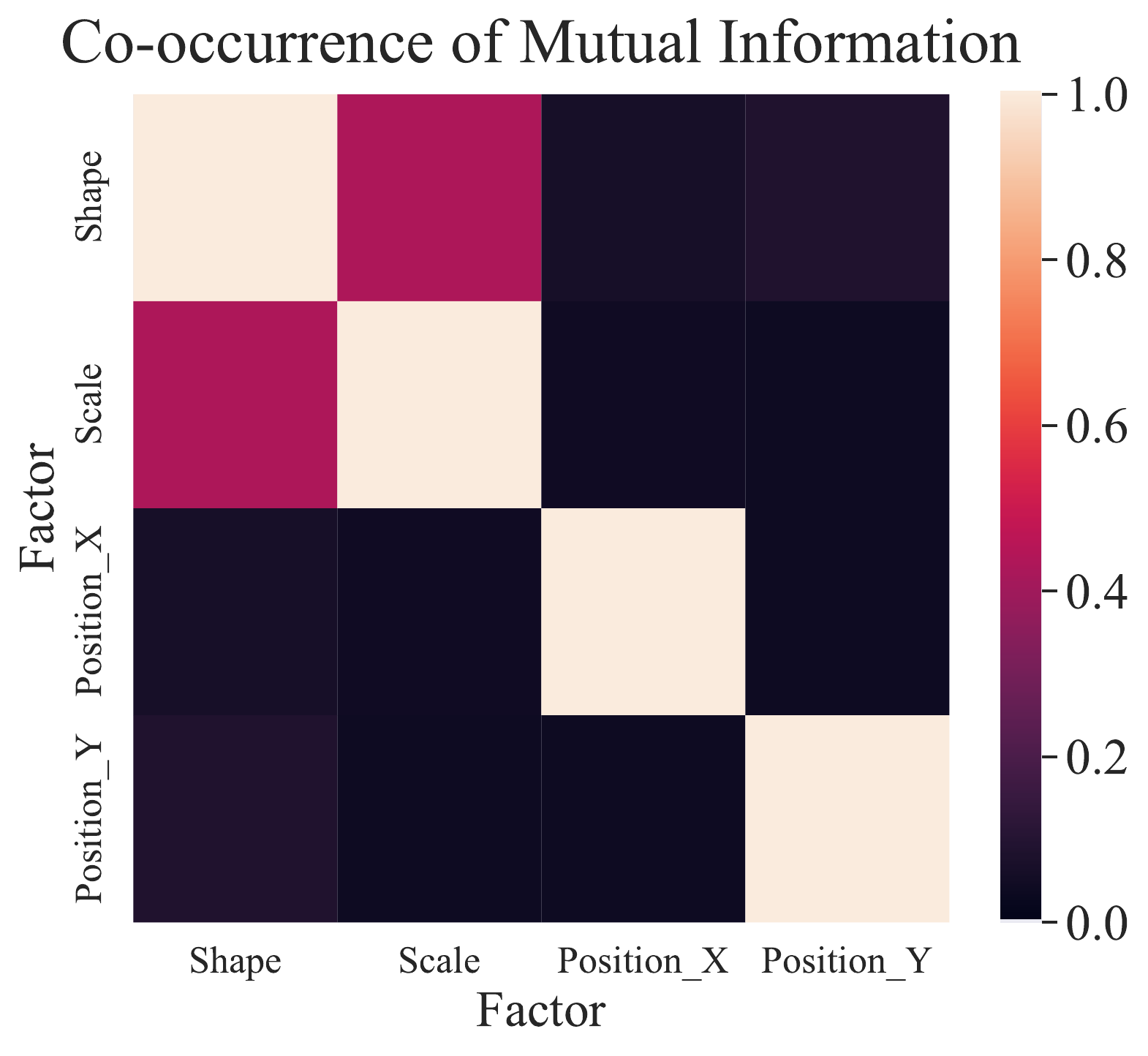}
    % \revfig{\includegraphics[width=0.3\textwidth]{dsprites_pk_co_with_orientation.pdf}}
  \end{center}
  \caption{The visualization of normalized co-occurrence of mutual information on dSprites.}
  \label{fig:cooccurrence}
  \vspace{-2em}
\end{wrapfigure}
To understand the disentanglement of a representation model, a basic question is how the representation dimensions correspond to data factors.  
After encoding an input image to a representation vector, we compute the normalized mutual information (MI) between each ground truth factor and each representation dimension, i.e. $R$ in \newmet{} defined as in Equation~\ref{eq:importance_mat}, to measure the correspondence. 

The mutual information between dimensions and the factors is included in Figure~\ref{fig:heatmaps}. The heatmaps are actually the transpose of the importance matrix $R^\top$. 
As shown by the brightness of the entries, the informativeness of the latent dimensions varies greatly. In fact, we can identify three types of columns: columns with \emph{single}, \emph{multiple} and \emph{no} bright elements, corresponding to three types of latent dimensions: \emph{disentangled} dimensions, \emph{entangled} dimensions and \emph{uninformative} dimensions. 
The \emph{disentangled} dimensions only capture information of one factor while the \emph{entangled} dimensions encode multiple factors. 
In contrast, the \emph{uninformative} dimensions cannot represent factors independently. 
As there are still \emph{entangled} and \emph{uninformative} dimensions, the representations are not fully disentangled. 
However, by extracting the subset of \emph{disentangled} dimensions we can derive a well-disentangled subspace. 

First, we analyze the model on dSprites dataset~\cite{dsprites17}. dSprites has five factors (shape, scale, orientation, position\_x, and position\_y).
But the orientation is ill-defined with ambiguity. 
For example, it is impossible to distinguish if a square rotates 0 degrees or 180 degrees. 
As shown in Figure~\ref{fig:separation}, the \emph{disentangled} dimensions occupy a significant proportion, indicating an evident partially disentangled pattern. 
% We note that this pattern can be also found on other datasets. 
We note that the degree of disentanglement varies on different datasets. An example  on Cars3D is shown in Figure~\ref{fig:heat_map_cars3d}. 
Cars3D is a dataset with 183 different car objects rendered from 4 elevations and 24 azimuths and the ground truth factors are not fully independent of Cars3D (see Appendix \ref{sec:more_qualitative}). It is extremely hard to represent its azimuth and the type of cars with few dimensions. Thus its object-type row and azimuth row in Figure~\ref{fig:heat_map_cars3d} are more spread out among multiple latent dimensions. 
% From the heatmap alone, we can tell it still shows some level of disentanglement but not as significant as that in dSprites. 
This also shows the difficulty of understanding the disentanglement of the high-dimensional representation model on complicated datasets. Therefore, we will continue to conduct a quantitative evaluation with our proposed MED metric in Section \ref{sec:quantitative_eval}. Further quantitative studies on more datasets is available in Appendix \ref{sec:more_qualitative}.

\subsection{Uniqueness of Factor-Representation Correspondence}
% In this section we investigate the properties of the well-disentangled subspace. 
In the ideal pattern of disentanglement, a representation dimension should uniquely correspond to only one factor. Now we show to what extent multiple factors are responded to by a single representation dimension in a well-disentangled subspace.
%, we study the normalized co-occurrence of mutual information.
Given the mutual information between the representation and the $i^{th}$ factor, noted as $I_i$, i.e., the $i^{th}$ row in the top-k version of Figure~\ref{fig:separation}, a good indicator of the uniqueness of factor-representation correspondence is the normalized co-occurrence of mutual information between the $i_1^{th}$ factor and the $i_2^{th}$ factor, which is defined as
\vspace{0cm}
\begin{equation}
    \widehat{C}_{i_1,i_2} = \frac{\langle I_{i_1} , I_{i_2} \rangle}{||I_{i_1}||_2 \cdot ||I_{i_2}||_2} = \frac{\sum_{d=0}^{K\cdot k-1} I(\tilde{\boldsymbol{c}}_d, \boldsymbol{v}_{i_1}) I(\tilde{\boldsymbol{c}}_d, \boldsymbol{v}_{i_2}) }{||I_{i_1}||_2 \cdot ||I_{i_2}||_2}, 
\end{equation}
where $\tilde{\boldsymbol{c}}$ is the representation after the selection process in Section~\ref{sec:partial_med}. 
We visualize the normalized co-occurrence of mutual information among the four factors by the learned representation in Figure~\ref{fig:cooccurrence}. 
It agrees that a dimension usually encodes only one factor.
Moreover, it indicates that the learned representation tends to encode shape and scale together, which also agrees with the intuitive analysis of the independence of factor pairs. For example, the shape and scale of dSprites objects are not disentangled and independent because objects with the same scale value but in different shapes have different pixel area. 

\begin{figure*}[ht]
    \centering
    \includegraphics[width=1.0\textwidth]{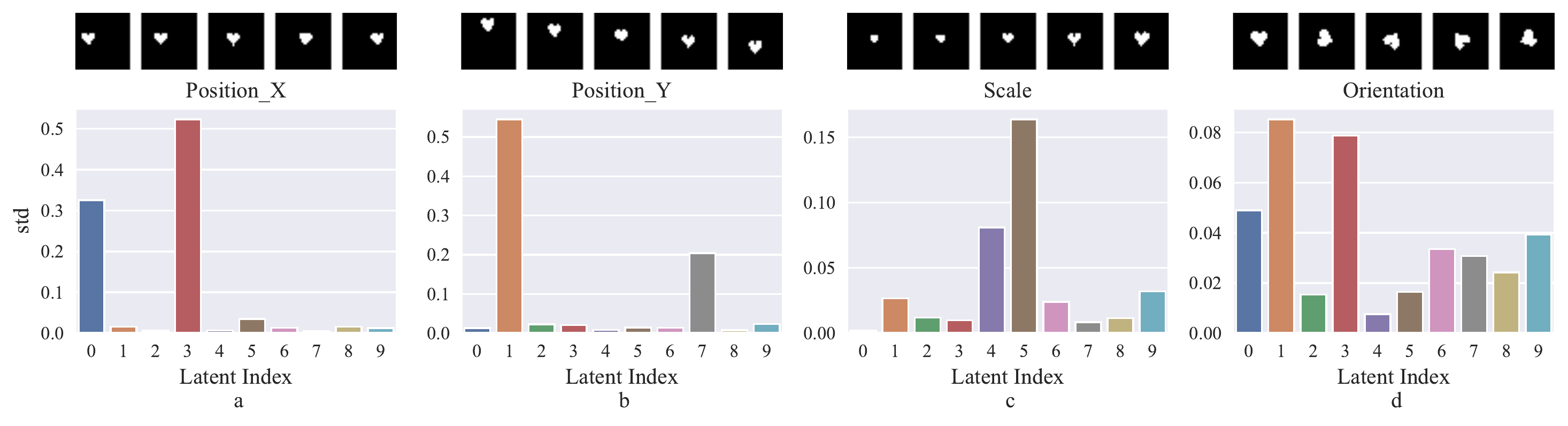}
    \caption{Representation variation when manipulating one factor only in the dimension-reduced version. In (a), (b), (c), \textit{position\_x}, \textit{position\_y} and \textit{scale} are manipulated respectively and only cause one dimension significantly variate. While, in (d), when manipulating the ill-defined factor \textit{orientation}, multiple dimensions variate.}
    \label{fig:factor_std}
\end{figure*}

\subsection{Influence by Manipulating Factors}
Another intuitive direction to study the relationship between representations and factors is the influence on representations when manipulating the factors. Given that the original representation vector dimension is much higher than the number of factors, we first make the representation more compact to have a more concise illustration.
Here, we reduce the representation dimension by the selection process of top-k \newmet{}  described in Section \ref{sec:partial_med}. 

Figure~\ref{fig:factor_std} shows the result of representation vector variation when changing only one factor at once. We take BYOL on dSprites as a representative. Here we set $k=2$ for top-k \newmet{}, i.e., we pick the 2 most disentangled dimensions for each factor and derive a 10-dim representation. 
We sample a set of images regarding a specific data factor. These images have all possible values of this data factor while having the same value for all other data factors.  We get 10-d representations from these generated images by a shared model.
Then, we compute the variance of each of the 10 dimensions across the images, leading to 10 scalars. The larger the variance is, the more that dimension responds to the factor. Figure~\ref{fig:factor_std}(a), (b) and (c) show how the reduced representation vector changes when manipulating the \textit{position\_x}, \textit{position\_y}, and \textit{scale} factors respectively. 
Note that we set $k=2$, therefore we see good disentanglement, with only exactly two representation dimensions having high variation. 
However, in Figure~\ref{fig:factor_std}(d)  we show a failure mode of the ill-defined factor \textit{orientation} that the change of factor causes multiple dimensions of reduced representation to have large variations, indicating that this factor is represented in an entangled way. 
From the results, we observe that manipulating different well-defined independent factors causes evident variance on disjoint sets of dimensions. Further, these results demonstrate the existence of a well-disentangled subset of latent dimensions. 
We also conduct a similar qualitative study where the representation dimension is reduced by the unsupervised PCA technique and observe a similar pattern. The details are provided in Appendix~\ref{sec:appendix_mani_factors}.

\section{Quantitative Evaluation}
\label{sec:quantitative_eval}
In this section, we conduct quantitative studies on the disentanglement property of contrastive learning methods. We first introduce the experiment setup in Section~\ref{sec:exp_setup}. 
Then we show quantitative results under both MED and existing disentanglement metrics in Section~\ref{sec:metric_disagree} which shows the disagreement of existing metrics to support the necessity of proposing MED. 
Finally, we make a full quantitative benchmark of methods of interest with MED in Section~\ref{sec:benchmark} and ablation study about the dimension of the representation model in Section~\ref{sec:dim_ablation}. \revtext{More quantitative studies are available in Appendix~\ref{sec:appendix_more_quan}.}

\subsection{Experiments Setup}
\label{sec:exp_setup}
The details for reproducibility are introduced in Appendix~\ref{sec:reproducibility}. Here we provide a brief description of the setup of the experiments.

\textbf{Datasets.} Representation disentanglement is usually evaluated on synthetic datasets, such as dSprites~\citep{dsprites17}, Cars3D~\citep{reed2015deep}, Shapes3D~\citep{3dshapes18}, and SmallNORB~\citep{LeCun2004LearningMF}. Besides those datasets, we also include a real-world dataset CelebA~\citep{liu2015deep}. CelebA contains human face images with 40 binary attributes. The attributes include fine-grained properties of the human face, such as whether wearing glasses or having wavy hair. 
%Table~\ref{tabl:dataset_factors} in Appendix explains the dataset factor details.
We include the details of dataset factors in the appendix.

\textbf{Evaluation Protocol.} 
We conduct experiments with both MED and the existing metrics to reveal the disagreement between existing metrics. Then we use MED as the main metric to study the disentanglement of contrastive learning methods. The implementation of evaluation metrics is adapted from the protocol provided by DisLib~\citep{locatello2019challenging}. All results are calculated with three random seeds and we report both the average score and the standard deviation. More details are introduced in Appendix~\ref{sec:appendix_metrics}.

\textbf{Reference Methods.} We investigate most of the popular disentangled representation learning methods as studied in the standard benchmark of DisLib~\cite{locatello2019challenging}. Besides, we also compare with a recently proposed ICA method called ICE-BeeM~\citep{icebeem}. Since we do not assume the ground truth factors are known during the training, we use its unconditional version. We term it EBM (energy-based model). For the contrastive learning methods, we evaluate not only negative-free methods such as BYOL~\cite{richemond2020byol}, Barlow Twins~\cite{zbontar2021barlow} and SimSiam~\cite{simsiam}, but also those using negative samples such as MoCo and MoCov2~\cite{he2020momentum}.

\textbf{Model Implementation.}
All methods use a shared architecture of encoder network as explained in the appendix. The latent dimension of contrastive learning methods is set to 1000 since they require a high-dimensional latent space to work. For other methods, the latent dimension is set to be 10 on synthetic datasets as in DisLib and 128 on CelebA dataset when evaluating with MED. For the evaluation with Top-k MED, the dimension of all methods is set to be 1000-d for fairness.
On dSprites, Cars3D and SmallNORB, we acquire checkpoints from DisLib if they are provided. We train our checkpoints on CelebA and Shapes3D. 

\begin{figure}
    \centering
    \includegraphics[width=\linewidth]{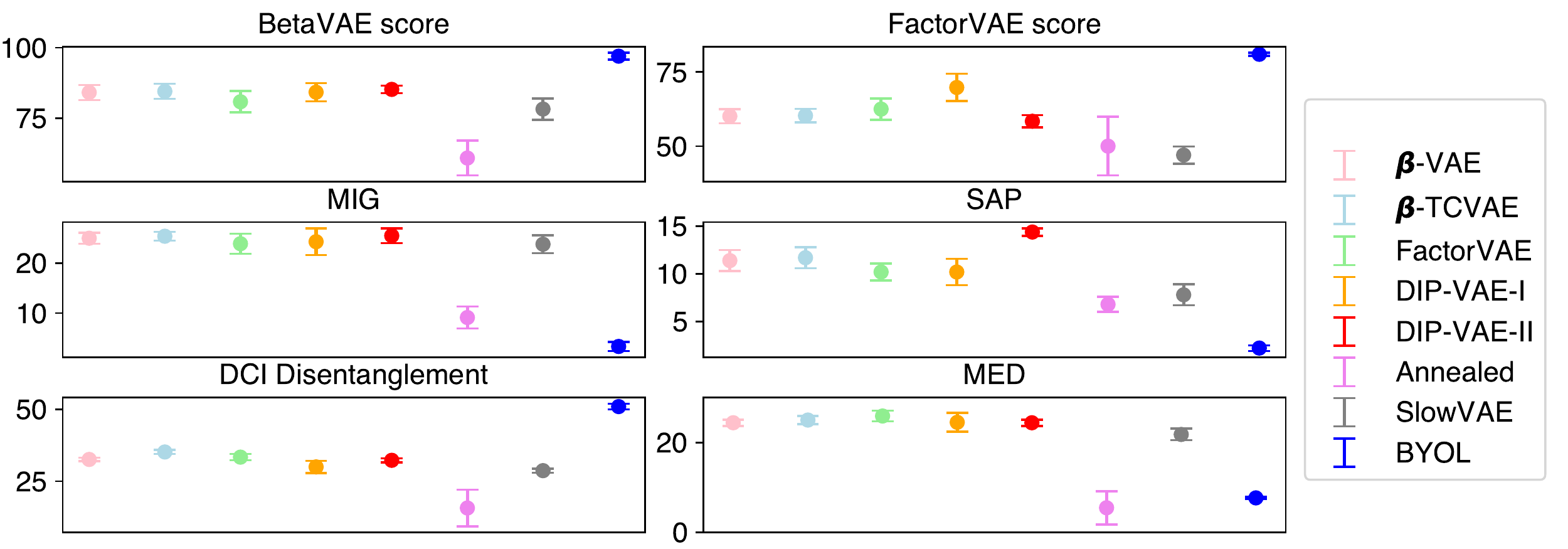}
    \caption{Evaluation with multiple metrics on SmallNORB. Y-axis is the corresponding disentanglement scores. The result shows the disagreement among existing metrics in high-dimensional space (BYOL).}
    \label{fig:disagreement}
    \vspace{-2em}
\end{figure}

\subsection{Disagreement of Existing Metrics}
\label{sec:metric_disagree}
As there is not a uniform definition of the ``disentanglement'', existing disentanglement metrics are motivated by different desired properties of a disentangled representation. These metrics had previously shown good agreement in the large-scale experiments of DisLib~\cite{locatello2019challenging}. However, when we extend the disentanglement study beyond low-dimensional scenarios, these metrics disagree significantly. 
We select representative low-dimensional VAE-based methods and a high-dimensional BYOL model to evaluate on a representative dataset, the SmallNORB dataset. The results are in Figure~\ref{fig:disagreement}, which show a significant disagreement among metrics on BYOL, while they agree on the low-dimensional VAE methods. 
Aligned with our analysis in Section~\ref{sec:med}, BetaVAE score, FactorVAE score, and DCI overestimate the disentanglement degree of the high-dimensional model while MIG and SAP underestimate it. 
%This explains why we need a new metric if we want to evaluate the disentanglement of contrastive learning. 
Since existing metrics fail to evaluate high-dimensional models, we opt to use MED as the main evaluation metric in the following sections. 
\revtext{Please refer to Appendix~\ref{sec:appendix_more_quan} and ~\ref{sec:disagreement_appendix} for more evidence of the disagreement of existing metrics. Moreover, in Appendix~\ref{sec:superority_of_med}, we provide the analysis by constructing scenarios where MED can still output the result aligned with human intuition while existing metrics fail to do meaningful measurement of disentanglement degree.}

\definecolor{gainsboro}{rgb}{0.86, 0.86, 0.86}
\definecolor{azure(web)(azuremist)}{rgb}{0.94, 1.0, 1.0}
\begin{table*}[!htp]
\centering
\small
\caption{MED and Top-k MED scores on multiple datasets. Methods in \textcolor{gray}{gray} are contrastive self-supervised learning methods. The latent dimension of VAE methods with * is 1000.}
\begin{tabular}{c|l|r|r|r|r|r} 
     \toprule
     Metrics & Model & dSprites & Shapes3D & Cars3D & SmallNORB & CelebA\\ 
     \midrule 
     \multirow{12}{*}{\textbf{MED}} & $\beta$-VAE & 32.6 (10.0)& 52.5 (9.4)  & 29.0 (2.2)&  24.4 (0.7) & 3.3 (0.5)\\
     & $\beta$-TCVAE   & 31.8 (7.4) & 53.2 (4.9) & \textbf{33.0 (3.8)} & 25.0 (0.9) & 4.7 (0.1)\\
     & FactorVAE  & 32.5 (10.1) & 55.9 (8.0) & 29.1 (3.0) & \textbf{25.9 (1.2)} & 0.6 (0.6)\\
     & DIP-VAE-I  & 18.8 (5.6) & 43.5 (3.7) & 19.4 (3.3) & 24.5 (2.1) & 3.7 (0.2)\\
     & DIP-VAE-II& 14.7 (5.5) & 52.6 (5.2) & 16.7 (4.1) & 24.4 (0.6) & --\\
     & AnnealedVAE & \textbf{35.8 (0.8)} & \textbf{56.1 (1.5)} & 15.5 (2.5) & 5.5 (3.7) & --\\
     & EBM & 6.8 (4.0) & 2.1 (2.6) & -- & 2.3 (1.7) & -- \\
    \cline{2-7} 
     & \CC{100} MoCo & \CC{100} 4.2 (0.5) & \CC{100} 6.1 (0.1) & \CC{100} 8.6 (0.4) & \CC{100} 4.9 (0.1) & \CC{100} \textbf{5.8 (0.1)} \\
     & \CC{100} MoCov2 & \CC{100} 3.5 (1.4) & \CC{100} 4.2 (0.4) & \CC{100} 6.5 (0.4) & \CC{100} 3.3 (0.2) & \CC{100} 4.8 (0.2) \\
     & \CC{100} BarlowTwins & \CC{100}6.0 (0.3) & \CC{100} 6.4 (0.3)  & \CC{100} 5.6 (1.4) & \CC{100}6.1 (0.1) & \CC{100} \CC{100} 4.2 (0.2) \\ 
     & \CC{100} SimSiam & \CC{100} 26.5 (0.1) & \CC{100} 12.3 (0.9) & \CC{100} 10.4 (0.1) & \CC{100} 10.7 (1.1)  & \CC{100} 5.3 (0.4)\\
     & \CC{100} BYOL  & \CC{100} 31.3 (0.4)  & \CC{100} 6.0 (0.5)& \CC{100} 9.7 (0.5) & \CC{100} 7.7 (0.2) & \CC{100} 4.8 (0.4)\\
     \cmidrule(r){1-7}
     \multirow{9}{*}{\makecell{\textbf{Top-k} \textbf{MED}}}  & $\beta$-VAE*     & 16.6 (6.2) & 19.2 (1.4) & 29.2 (2.0) & 15.8 (2.1) & 4.5 (0.3)\\
 	& $\beta$-TCVAE*   & 11.2 (0.4)        & 25.0 (0.5)      & 20.0 (1.8) & 23.0 (0.6)  & 3.6 (0.2)\\
	& FactorVAE*       & 3.2 (3.8)  & 8.2 (4.0)         & 7.8 (1.6)  & 4.8 (1.0)   & 5.0 (0.3)\\
	& DIP-VAE-I*       & 7.0 (1.3)  & 16.2 (0.9)         & 24.6 (2.2) & 20.9 (2.7)  &2.5 (0.9)\\
	 \cline{2-7}
     &  \CC{100} MoCo &  \CC{100} 16.1 (2.0) & \CC{100}18.1 (0.6) & \CC{100} 26.6 (1.6) & \CC{100} 17.9 (0.8) & \CC{100} \textbf{7.9 (0.1)} \\
     &  \CC{100} MoCov2 & \CC{100} 14.7 (1.0) & \CC{100}13.6 (1.7) & \CC{100} 24.5 (2.1) & \CC{100} 15.1 (0.9) & \CC{100} 6.6 (0.7) \\
     &  \CC{100} BarlowTwins & \CC{100} 21.7 (1.3) & \CC{100}20.0 (0.3)  & \CC{100} 23.8 (2.5) & \CC{100} 24.5 (1.5) &\CC{100} 5.7 (0.2) \\
     &  \CC{100} SimSiam &  \CC{100} 39.1 (0.4) & \CC{100} \textbf{30.0 (2.0)} & \CC{100} \textbf{32.7 (2.3)} & \CC{100} \textbf{28.4 (1.9)} & \CC{100} 7.2 (0.6)\\
     &  \CC{100}  BYOL  & \CC{100} \textbf{53.7 (0.7)} & \CC{100} 19.7 (1.3) & \CC{100} 31.8 (1.3) & \CC{100} 25.7 (0.3) & \CC{100} 6.8 (0.7)\\
     \bottomrule
\end{tabular}
\label{table:benchmarks}
\end{table*}

\subsection{Disentanglement Benchmark with Contrastive Learning Methods}
\label{sec:benchmark}
For the benchmarking of disentanglement, we use both MED and the partial version of MED, i.e. Top-k MED. For Top-k MED, we set the MED partial evaluation hyperparameter $k=2$ for dSprites, Shape3D, and SmallNORB, and $k=3$ for Cars3D and CelebA. The values of $k$ are chosen such that the selected dimensions are roughly close to the latent space dimension of the low-dimensional reference methods. And we extend the latent dimension of all methods to 1000 when evaluating Top-k MED for fairness. We note that despite our hyperparameter search, we were unable to train good EBM weights on Cars3D and CelebA, so we keep that section empty. The results are in Table~\ref{table:benchmarks}. \revtext{We also encourage readers to read the results of existing metrics in Table~\ref{table:full_results} in Appendix~\ref{sec:appendix_more_quan}.}

\begin{figure}
    \centering
    \revfig{\includegraphics[width=.9\textwidth,keepaspectratio]{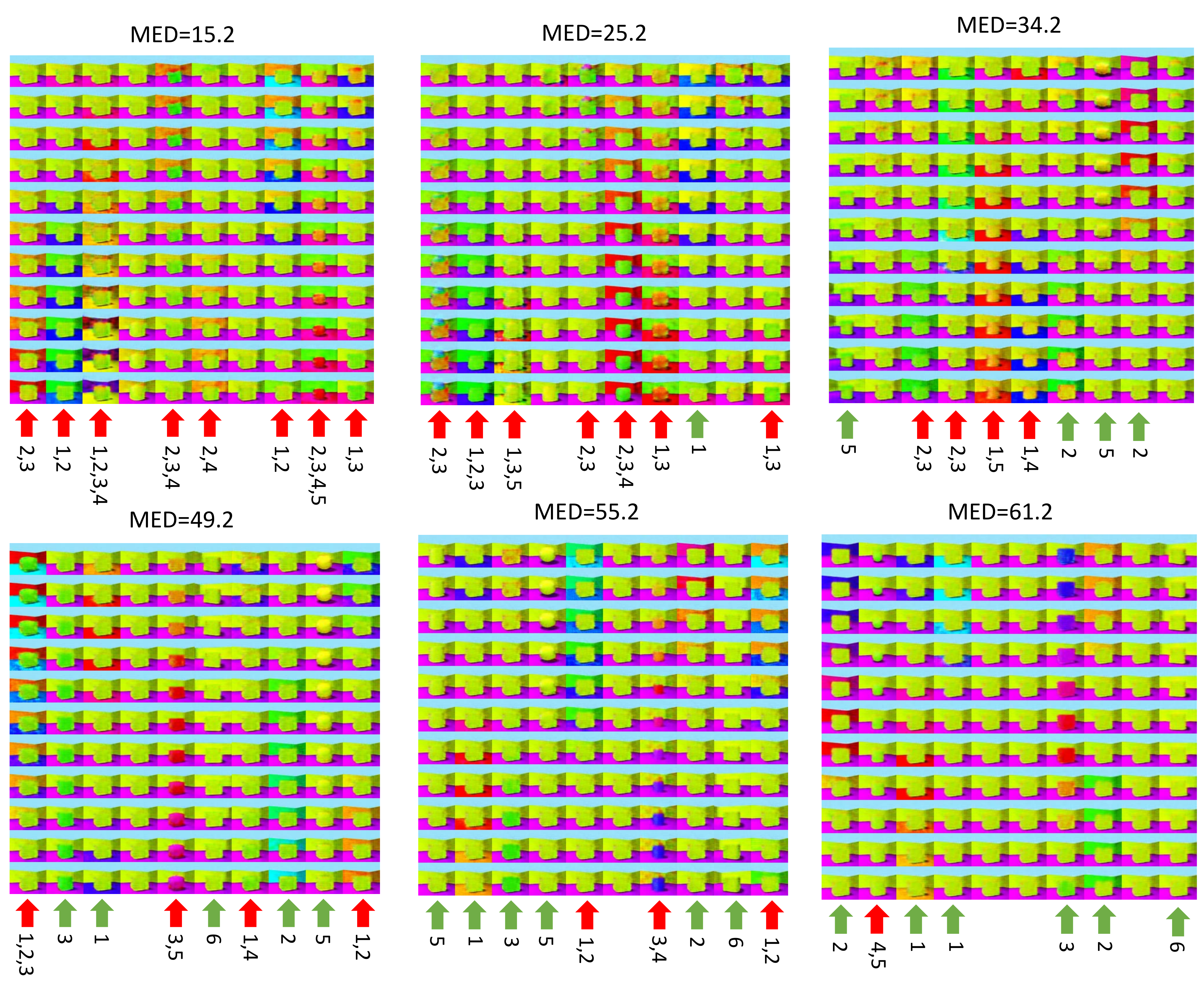}}
    \caption{\revtext{The visualization of latent traversing by VAE models from~\citenst{van2019disentangled}. Each subfigure is the visualization for one model, annotated with the MED score of that model. The involved data factors include (1) Floor color; (2) Wall color; (3) Object color; (4) Object scale; (5) Object shape and (6) Orientation. For each column in the subfigure, the value of only one latent dimension changes, i.e. latent traversal for that dimension. Below each column, we also annotate the actual changing factors in this column, obtained with human manual inspection. If a model is fully disentangled, we expect each column to correspond to only one ground truth factor. The visualization shows that if a model achieves a higher MED score, it has more dimensions disentangled and responsive to only one factor while fewer dimensions entangled to multiple factors.}}
    \label{fig:med_traverse}
    \vspace{-2em}
\end{figure}

In the upper part of Table~\ref{table:benchmarks}, we show the MED score for previous disentangled methods as well as contrastive methods. We find that contrastive methods achieve significantly lower disentanglement scores on 3 of the 5 datasets (Shapes3D, Cars3D, SmallNORB). Contrastive methods achieve slightly higher disentanglement scores on CelebA. On dSprites, some negative-free contrastive methods (SimSiam, BYOL) achieve scores close to SOTA, but the other contrastive methods' score is much lower. In summary, in most cases, contrastive methods have inferior disentanglement properties compared to the best methods; only in a few settings do contrastive methods achieves scores comparable to SOTA scores. 

These results are disappointing but not surprising, since contrastive methods are not explicitly designed to maximize the feature disentanglement. 
Further, since the underlying number of factors is usually quite small, on the order of 10, the 1000-d feature space will likely have dimensions that are either not related to the ground truth factors, or capture a combination of the ground truth factors. This result does not contradict with \citet{zimmermann2021contrastive} since (1) they find that the learned feature is a linear transformation of the ground truth factors, which doesn't necessarily disentangle and (2) they use augmentations on factors that cannot be done in practice. 

The lower part of Table \ref{table:benchmarks} shows top-k MED measurements on various methods. We find that contrastive learning methods (especially the negative-free ones) in general show a better disentanglement in a selected subspace and the disentanglement is stronger than the reference methods. 
This shows that there exists a subspace in the learned representation that is well disentangled. 
Moreover, when we compare the subspace in contrastive methods (gray part in the lower section of Table~\ref{sec:benchmark}) to the traditional approach that directly trains a low-dimensional latent space (non-gray part in the upper section of Table~\ref{sec:benchmark}), we find that the disentanglement of the former is usually better than or on par with that of the latter. 
This means that we probably should not constraint the dimension of the latent space and require it to be fully disentangled, but rather should encourage to use high dimensional latent spaces and only require it to have a subset with good disentanglement properties. 

To conclude, we find that the high-dimensional contrastive methods, including negative-free ones, do not learn a fully disentangled representation. However, there exists a subspace in the learned representation that is well disentangled. Such a subspace can show much better disentanglement property than previous SOTA approaches. Despite the fact that these methods require a high dimension to train, such a subspace can serve as proxy between contrastive learning methods and a more compact and disentangled low-dimensional representation.

\subsection{\revtext{Latent Traverse on Shapes3D}}
Now, we provide visualizations to show that MED can provide evaluation results aligned with human intuition about the disentanglement degree regarding data factors.
Because CL methods are not generative models and have high dimensions, they are hard to be adopted for visualization with latent traversal. 
We thus adopt VAE-based generative models here to perform latent traverse.
We follow the practice in DisLib~\cite{locatello2019challenging} to use Shapes3D for the latent traverse. 
For fairness and reproducibility, we trained VAE models on Shapes3D with the published configurations provided by~\citenst{van2019disentangled} because DisLib~\cite{locatello2019challenging}.
The results are shown in Figure~\ref{fig:med_traverse}. 
For each column in each subfigure, only the value of one dimension of the latent code is manipulated. 
The manipulation is performed the same way as the default setup for traverse visualization in DisLib. 
Given the six factors on Shapes3D with index, we indicate under each subfigure if 
(1) \textcolor{red}{red arrow}: a dimension is entangled to more than one factor or 
(2) \textcolor{green}{green arrow}: a dimension is disentangled and responsive to only one factor. 
Through the visualizations and the corresponding MED scores, we can clearly see that MED can well represent the disentanglement degree. 
We could observation a clear pattern that model with higher MED score has more disentangled representations.
This demonstrates that the results from MED scores are aligned with the intuition of humans.

\begin{figure}
\centering
\begin{subfigure}{.44\textwidth}
  \centering
  \includegraphics[width=\linewidth]{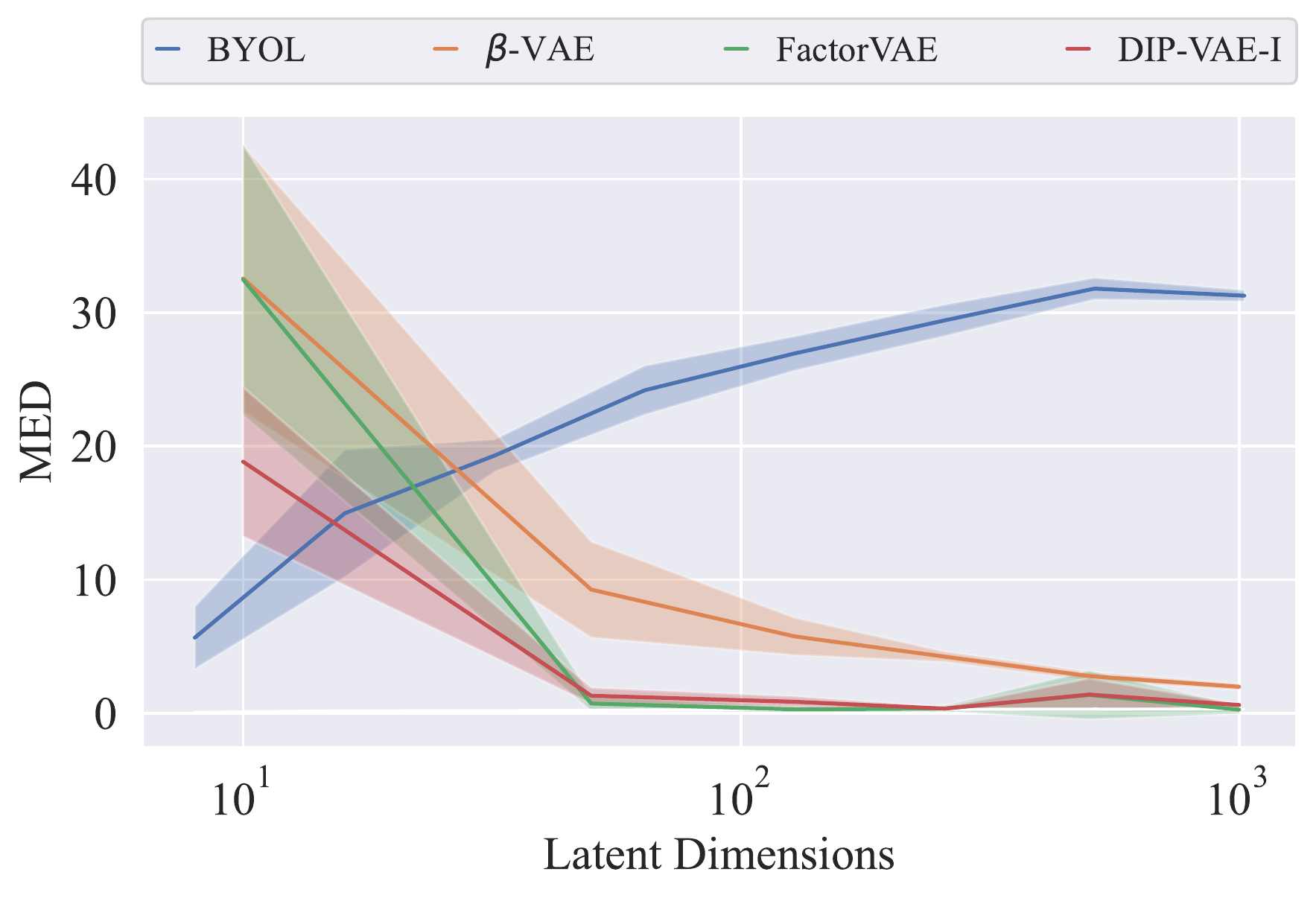}
  \caption{MED score w/ dimension change.}
  \label{fig:dimension_ablation_med}
\end{subfigure}%
\hfill
\begin{subfigure}{.44\textwidth}
  \centering
  \includegraphics[width=\linewidth]{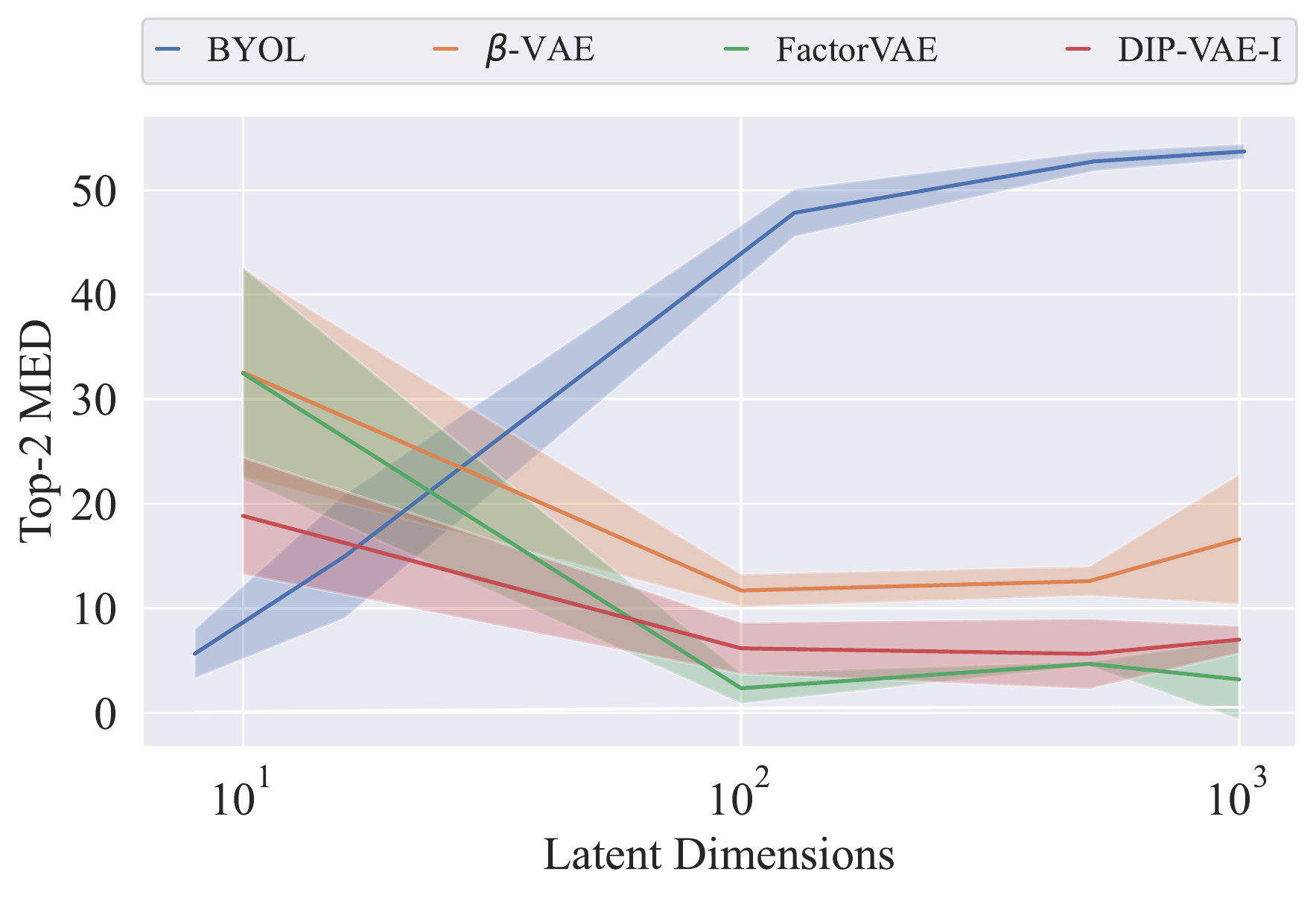}
  \caption{Top-2 MED score w/ dimension change.}
  \label{fig:dimension_ablation_med_top2}
\end{subfigure}
\caption{The influence of the representation dimension on the (a) MED score and (b) Top-2 MED score on dSprites dataset. The disentanglement property regarding both metrics is enhanced with increasing representation dimension for the BYOL method but decreased for all tested VAE-based methods. The minimum dimension is 10 for VAE-based methods, and 8 for BYOL.}
\label{fig:dimension_ablation}
\vspace{-2em}
\end{figure}

\subsection{Influence of Dimension}
\label{sec:dim_ablation}
As representation dimension is found as a core variable in disentangled representation learning, we evaluate the influence of dimension over disentanglement by both MED and top-k MED on dSprites with BYOL as an example versus VAE-based methods. The results are shown in Figure~\ref{fig:dimension_ablation}. 
We find that the BYOL's MED score and top-2 MED score increase along with the latent dimension. The scores plateau at around 512 dimensions. This is consistent with previous literature on the difficulty to train informative contrastive models with low latent dimensions~\cite{grill2020bootstrap}. We further show that a lower latent dimension leads to a less disentangled subspace as well. On the contrary, we also note that the VAE methods fail to scale to higher latent dimensions. 
We find a large gap between higher-dimensional VAEs and their 10-dim versions. This suggests the gap between existing disentangled representation methods and the real-world data complexity, which can not be represented in a limited dimension.

\section{Conclusion}
In this paper, we provide an empirical study of the disentanglement property of contrastive learning without negatives for the first time. 
In the high-dimensional space, we find the difficulty of adopting the existing disentanglement metrics. Therefore, we propose a new metric, MED and top-k MED, to evaluate disentanglement based on mutual information. 
The evaluation shows that even without negative samples, contrastive learning can learn a well-disentangled subset of representation. 
Recently, the study of contrastive learning, or general self-supervised learning, is still motivated by empirical observations. We hope our work can reveal some clues to motivate future theoretical justifications.

\section*{Acknowledgement}
We appreciate the help from Jinhyung Park on paper writing. This work is supported by the Ministry of Science and Technology of the People's Republic of China, the 2030 Innovation Megaprojects "Program on New Generation Artificial Intelligence" (Grant No. 2021AAA0150000). This work is also supported by a grant from the Guoqiang Institute, Tsinghua University.

\medskip
\bibliography{neurips2022_conference}
\bibliographystyle{abbrvnat}

\newpage
\appendix
\section{Reproducibility}
\label{sec:reproducibility}
In this section, we provide the information required to reproduce our results reported in the main text. And we commit to making the code implementation and evaluating checkpoints public. Our experiments are run on a machine with an AMD Ryzen Threadripper 3970X 32-Core Processor and a GeForce RTX 3090 GPU.

\paragraph{VAE methods implementation.} For the results on synthetic datasets, i.e., dSprites, Cars3D, SmallNORB, and Shapes3D, the disentanglement score is from the original logs of DisLib~\cite{locatello2019challenging}~\footnote{https://github.com/google-research/disentanglement\_lib}. In the released logs, each method has different training configurations, and our reported result is from the configuration with the highest average performance overall the provided random seeds. For the evaluation on the CelebA dataset, we follow an open-sourced implementation in Pytorch~\footnote{https://github.com/AntixK/PyTorch-VAE} and align the encoder architecture of all methods to be the same as described in Appendix~\ref{sec:appendix_byol}. For the results on Shapes3D, because DisLib does not release the pretrained checkpoints, we use the same open-sourced implementation to reproduce with the configuration indicated by DisLib. Parameters are kept as the default well-tuned version in the provided implementation. When the latent dimension is 1000, training of BetaTC VAE will collapse with the default hyperparameters, we have to decrease the $\beta$ to 3.0 to work it around.

\paragraph{GAN methods implementation.} Limited by the text length, we do not include the performance of GAN methods in the main text, but we will report some in the following appendix content. It is hard to include GAN methods' performance in the benchmark as the training is not always stable and the discriminator weights are usually not provided in many public codebases. When evaluating the methods on synthetic datasets, the FactorVAE scores of InforGAN, IB-GAN, and InfoGAN-CR are provided in the paper of~\citenst{lin2020infogan}. But the evaluation of other metrics in~~\citenst{lin2020infogan} uses a not aligned settings with~\citenst{locatello2019challenging}, so we check its officially release~\footnote{https://github.com/fjxmlzn/InfoGAN-CR} to evaluate the provided implementation and model weights under the unified evaluation setup.  We perform the same evaluation process for results on the CelebA dataset.

\paragraph{Energy-based Model (EBM).} We refer to the implementation of ICE-BeeM~\citep{icebeem} for this method. We use the officially released codebase for it~\footnote{https://github.com/ilkhem/icebeem}. The encoder implementation has been aligned with our default already. The only modification we make is to use the unconditional version instead of its default conditional version in loss computation to satisfy the fully unsupervised settings. 

\paragraph{Contrastive Learning implementation.} Our implementations are based on the public and official implementations of MoCo/MoCov2~\footnote{https://github.com/facebookresearch/moco}, BYOL/SimSiam~\footnote{https://github.com/lucidrains/byol-pytorch} and Barlow Twins~\footnote{https://github.com/facebookresearch/barlowtwins}. The details of implementation are explained in Appendix~\ref{sec:appendix_byol}.

\paragraph{Evaluation Protocol.} 
For \newmet{}, we first compute MI following the implementation of MIG by DisLib~\cite{locatello2019challenging}. Then we calculate the entropy disentanglement score in the same way as the DCI Disentanglement score in DisLib. 
For other disentanglement metrics evaluation, we use the implementation of DisLib. 
The settings of some important parameters are provided in Appendix~\ref{sec:appendix_metrics}.

\subsection{Implementation of contrastive learning model}
\label{sec:appendix_byol}
\paragraph{Architecture.} 

%We describe the implementation details of BYOL in this part. 
To make a fair comparison with previous methods, we follow the encoder architecture in Factor VAE~\citep{kim2018disentangling}. The pipeline details are shown in Table~\ref{table:encoder}. After each convolutional layer in the figure, there is a ReLU activation layer and a group normalization (group number = 4) layer for BYOL. So, the encoder is a stack of (Conv-ReLU-GN) blocks. For other contrastive learning methods, we keep the default batch normalization to replace GN. By default, the final output channel number is 1000, i.e, $D=1000$. For other details of contrastive learning methods, we follow the convention in their official implementations.

Besides the representation network (encoder), BYOL also has a projector network and a predictor network. Both of them consist of a pipeline ``Linear $\longrightarrow$ BN $\longrightarrow$ ReLU $\longrightarrow$ Linear''. The projection dimension is 256, and the hidden dimension of the projector is 4096. The predictor keeps a 256-dimensional feature vector in its pipeline.

\renewcommand\arraystretch{1.1}
\small
\begin{table}[!htp]
\centering
\caption{The encoder architecture for our implemented contrastive learning methods on synthetic datasets.  Besides, there is a ReLU activation layer and a possible normalization layer following each convolutional layer to create a stack of (Conv-ReLU-Norm) blocks.}
\begin{tabular}{l} 
     \toprule
     \textbf{Encoder} \\
     \midrule
     \textbf{input}:  $64\times64$ images \\ 
     \textbf{pipeline}:  \\
     \quad \quad \quad 4$\times$4 conv, stride 2, 32-channel \\
     \quad \quad \quad 4$\times$4 conv, stride 2, 32-channel\\
     \quad \quad \quad 4$\times$4 conv, stride 2, 64-channel\\
     \quad \quad \quad 4$\times$4 conv, stride 2, 64-channel\\
     \quad \quad \quad 4$\times$4 conv, stride 2, 128-channel\\
     \quad \quad \quad 1$\times$1 conv, stride 1, $D$-channel\\
     \bottomrule
\end{tabular}
\label{table:encoder}
\end{table}

\paragraph{Training settings.} We make minor modifications to the training setting of default BYOL to apply to contrastive learning methods without negative samples. For training on all datasets, the images are resized to 64x64. For data preprocessing, we copy 1-channel images of dSprites and SmallNORB to be 3-channel. During the training stage, we use such a pipeline of augmentation (in \textit{PyTorch}-style):
\begin{enumerate}
    \item \textit{RandomApply(transforms.ColorJitter(0.8, 0.8, 0.8, 0.2), p=0.3)}
    \item \textit{RandomHorizontalFlip()}
    \item \textit{RandomApply(transforms.GaussianBlur((3,3), (1.0, 2.0)), p=0.2)}
    \item \textit{RandomResizeCrop(size=(64, 64), scale=(0.6,1.0))}
    \item normalization.
\end{enumerate}
For the normalization, the pixel value of images from dSprites and SmallNORB is uniformly normalized from [0,255] to [0,1.0]. For Cars3D, Shapes3D, and CelebA, we adopt the commonly used Imagenet-statistic normalization for preprocessing the RGB image pixel values.

During training, we use Adam optimizer by default, whose learning rate is $3e-4$ without weight decay. The batch size is set to 512 by default. For evaluation on dSprites, Shapes3D, and CelebA, we select the weights after training for 15 epochs for evaluation. We select the weights after training for 140 epochs for evaluation on Cars3D and the weights of the 200th epoch on SmallNORB considering the small scale of these two datasets.

To decrease the influence of randomness, we train each model configuration multiple times with different random seeds (seed=0, 1, 2). We report the average and standard deviation. To be precise, as our implementation  is based on Pytorch, we initialize the libraries of \textit{numpy}, \textit{torch}, \textit{torch.cuda}, and \textit{random} with the same random seeds. 

\begin{table}
\scriptsize
\centering
\caption{The factors on all the datasets we investigate the disentanglement on.}
\begin{tabular}{c|l|l|l|l|l} 
     \toprule
      & dSprites & Shapes3D & Cars3D & SmallNORB & CelebA \\
     \midrule
     \multirow{6}{*}{\textbf{\makecell{Factors \\ (\# of values)}}} 
     & Shape (3)   & Floor hue (10)  & Elevation  (4)& category (10) & 40 attributes \\
     & Scale  (6)  &  Wall hue (10) & Azimuth (24) & Elevation (9) & (2 for each) \\
     & Orientation  (40) & Object hue (10) &  Object id (183) & Azimuth (18)\\
     & Position X  (32)  & Scale  (8) & & Lighting (6)\\
     & Position Y  (32) & Orientation (15)& & \\
     & & Shape  (4) & &  \\
     \bottomrule
\end{tabular}
\label{tabl:dataset_factors}
\end{table}

\subsection{Evaluation Metrics}
\label{sec:appendix_metrics}
In the main text, we compare the evaluation metrics provided in the DisLib protocol with our proposed MED metric. Here we provide more details about them. Moreover, we would conduct evaluations under all of them in the next section.

\textbf{BetaVAE Metrics.} Introduced in~\citet{higgins2016beta}, BetaVAE score assumes each dimension corresponds to one category in a linear classifier. Representations are obtained after the generated samples with only one factor fixed. Then we calculate the summation of the divergence between different representations and put it into a linear classifier. The classifier is trained to predict the index $k$ for the fixed data factor. The accuracy of this linear model is the value of the BetaVAE score.

\textbf{FactorVAE Metrics.} ~\citet{kim2018disentangling} argues the BetaVAE score has the tendency to fail into a spurious disentanglement and proposes a new metric based on a majority vote classifier. Representations are obtained after the generated samples with only the $k$-th factor fixed. Normalizing each dimension in representations in terms of standard deviation. The index of dimension with the lowest variances of normalized representation and the factor index $k$ are the input and the output of the linear classifier. The accuracy of the classification is the FactorVAE score.

\textbf{Mutual Information Gap.}~\citet{chen2018isolating} assumes the disentanglement model has the property that most information of one specific factor is contained in one dimension or a group of certain dimensions. The mutual information gap is the summation of the difference between the highest and second-highest normalized mutual information between a fixed factor and the dimensions of the output representation vector. The formula can be illustrated below:
\begin{equation}
    \frac{1}{K}\sum_{k=1}^{K}\frac{1}{H_{z_k}}(I(v_{j_k}, z_k) - \max_{j \neq j_k}I(v_j, z_k)),
\end{equation}
where $K$ is the overall number of ground truth factors. $v$ is the latent representation and $z_k$ is the factors of latent variables and $j_k = \argmax_{j}I(v_j,z_k)$.

\begin{figure*}
    \centering 
    \includegraphics[width=\linewidth]{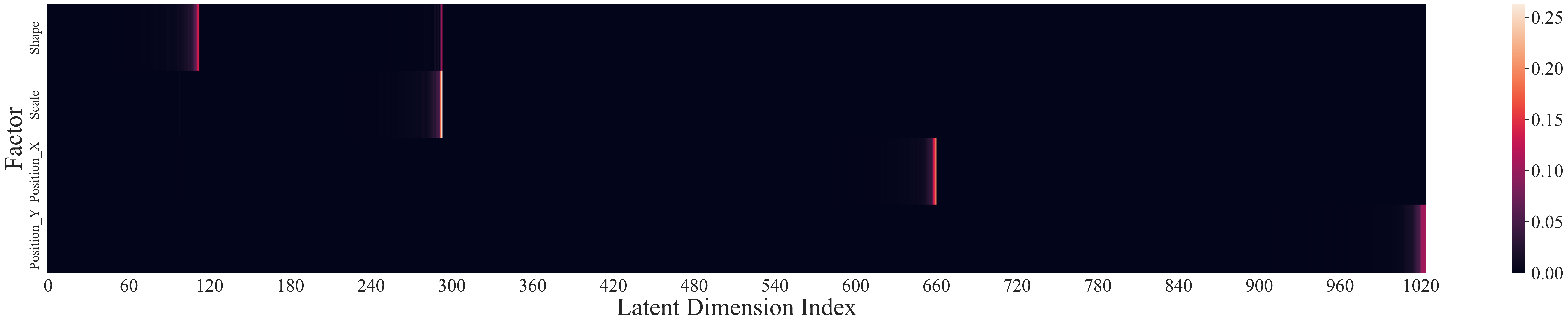}
    \caption{The importance distribution for the representation learned from BYOL on dSprites. Here, we follow the practice of DisLib to use a Gradient Boosting Tree (GBT) regressor to determine the importance matrix of each latent dimension in predicting each factor. Compared with the Mutual Information distribution shown in Figure~\ref{fig:separation}, the importance distribution is significantly more sparse. Sparsity is encouraged when constructing the GBT regressor. This makes it hard to study the true representation pattern.}
    \label{fig:importance_distri}
\end{figure*}

\textbf{DCI disentanglement.} As~\citet{eastwood2018framework} suggest, 
%Disentanglement, Completeness, and Informativeness are three properties of representation. 
the disentanglement can also be measured by the entropy of relative importance for each dimension in predicting factors. First, we have to know the importance of each dimension of the representation for predicting each factor. The importance is determined by a regressing model such as Lasso or Random Forest in the original DCI implementation~\citep{eastwood2018framework} or Gradient Boosting Tree in DisLib implementation~\citep{locatello2019challenging}. We note the importance matrix $R$ where $R_{ij}$ is the importance of the i-th dimension in prediction the j-th factor. Then the disentanglement score for the i-th dimension is defined as $D_i=(1-H_K(P_i))$ where $H_K(P_i) = -\sum_{k=0}^{K-1}P_{ik}log_KP_{ik}$ denotes the entropy and $P_{ij} = R_{ij}/\sum_{k=0}^{K-1}R_{ik}$ denotes the normalized importance of the i-th dimension in prediction the j-th factor. Finally the overall disentanglement score is calculated as $D=\sum_i \rho_i D_i$ where $\rho_i = \sum_j R_{ij} / \sum_{ij} R_{ij}$ is the weighting factor of the each dimension's informativeness in representing factors.

\textbf{SAP.}~\citet{kumar2017variational} proposes the Separated Attribute Predictability (SAP) score. SAP is computed with classification score of predicting $j^{th}$ factors on $i^{th}$ dimension as the $ij^{th}$ entry. SAP is the mean of the difference between the highest and second-highest scores for each column.

We follow the implementation provided by DisLib~\citep{locatello2019challenging} for the evaluation protocol. Despite exceptions, the evaluation batch size is 64, the \textit{prune\_dims.threshold} is 0.06. If a classifier is required to be trained during evaluation, \textit{num\_train} is 10000, and \textit{num\_eval} is 5000. For Mutual information computation, the discretizer function is the histogram discretizer, and the number of bins in the discretization is 20. 
For the evaluation of MIG and SAP on dSprites, SmallNORB, Cars3D, and Shapes3D, BYOL representation vectors are reduced to 10 dimensions by PCA to be aligned with other methods. For the evaluation of MIG and SAP on CelebA, to have a fair comparison, the representation vectors of all methods are reduced to 40 dimensions. For the implementation of our proposed MED, the basic logic is the same as DCI Disentanglement, but we replace the classifier output with the mutual information based scores.

\section{More Qualitative Study}
\label{sec:more_qualitative}
We provide more qualitative studies about the disentanglement property shown by the contrastive learning here. We still use BYOL as an example of the negative-free contrastive learning methods.

\begin{figure}
\begin{subfigure}{\textwidth}
  \centering
  \includegraphics[width=\linewidth]{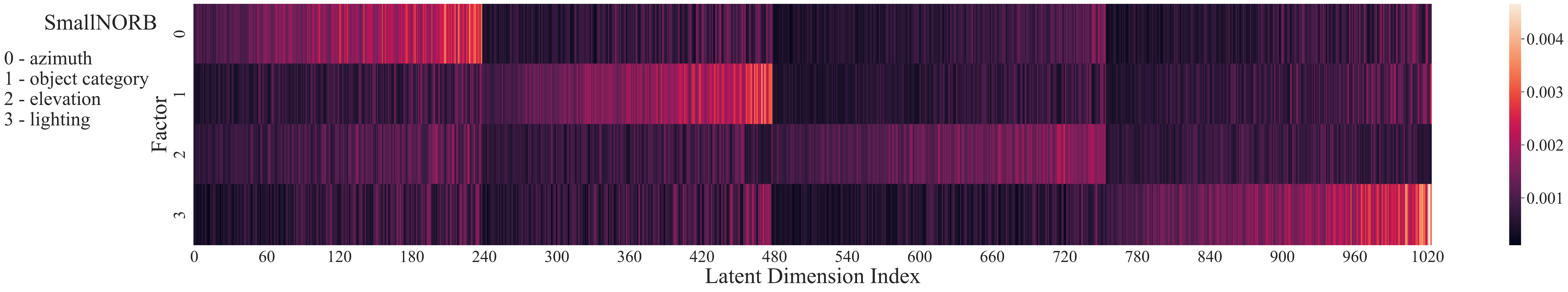}
  \caption{}
  \label{fig:mi_smallnorb}
\end{subfigure}
\\
\begin{subfigure}{\textwidth}
  \centering
  \includegraphics[width=\linewidth]{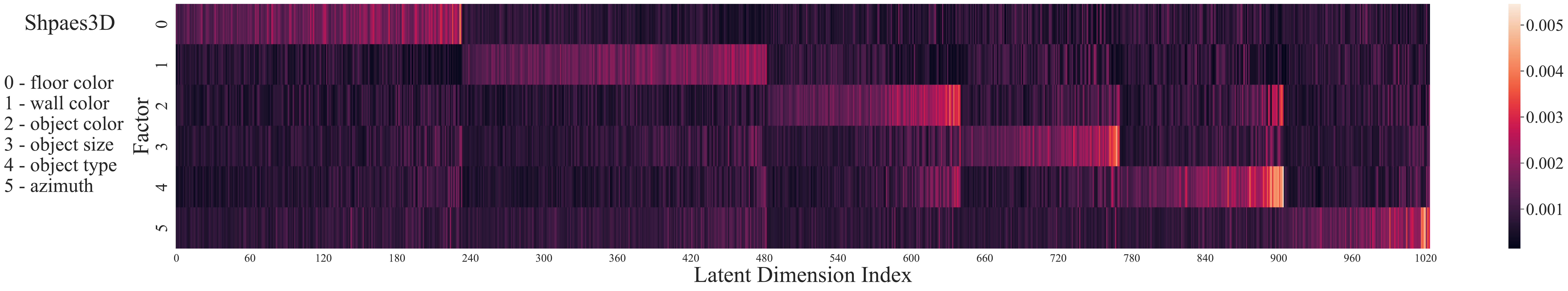}
  \caption{}
  \label{fig:mi_shapes3d}
\end{subfigure}
\begin{subfigure}{\textwidth}
  \hfill
  \revfig{\includegraphics[width=\linewidth]{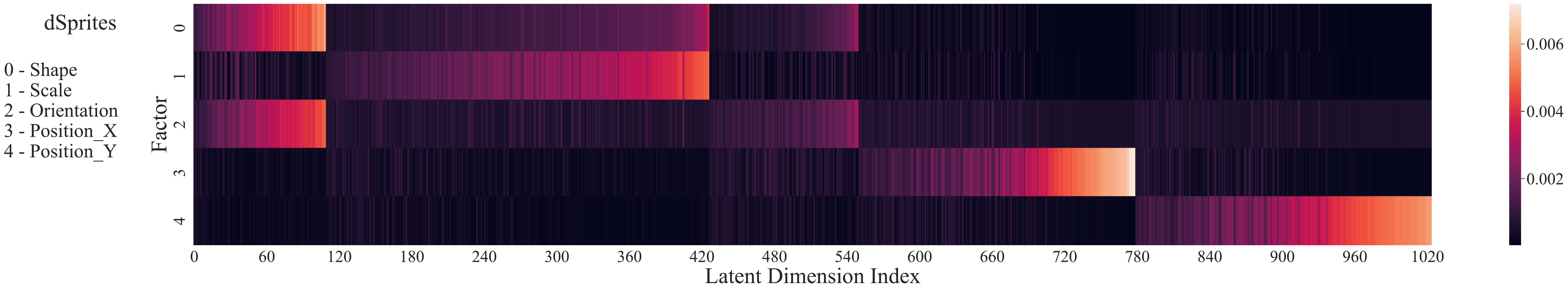}}
  \caption{}
  \label{fig:mi_dsprites_full}
\end{subfigure}
\caption{The mutual information distribution on SmallNORB(a),  Shapes3D(b) , \revtext{and dSprites including all factors(c). }}
\label{fig:mi_all}
\end{figure}

\begin{figure}
\begin{subfigure}{.22\textwidth}
  \centering
  \revfig{\includegraphics[width=\linewidth]{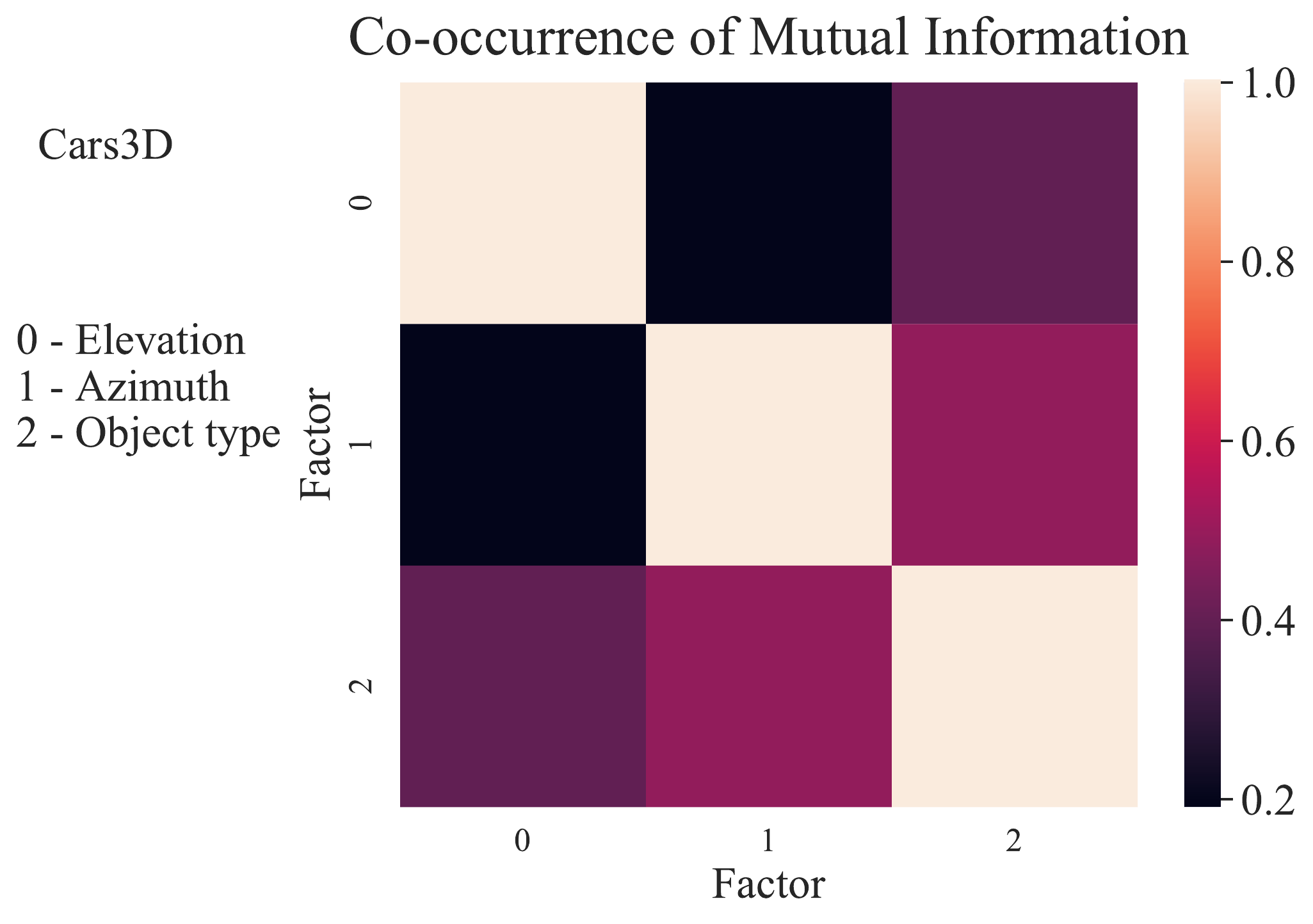}}
  \caption{}
  \label{fig:cooccurence_cars3d}
\end{subfigure}%
\begin{subfigure}{.22\textwidth}
  \centering
  \includegraphics[width=\linewidth]{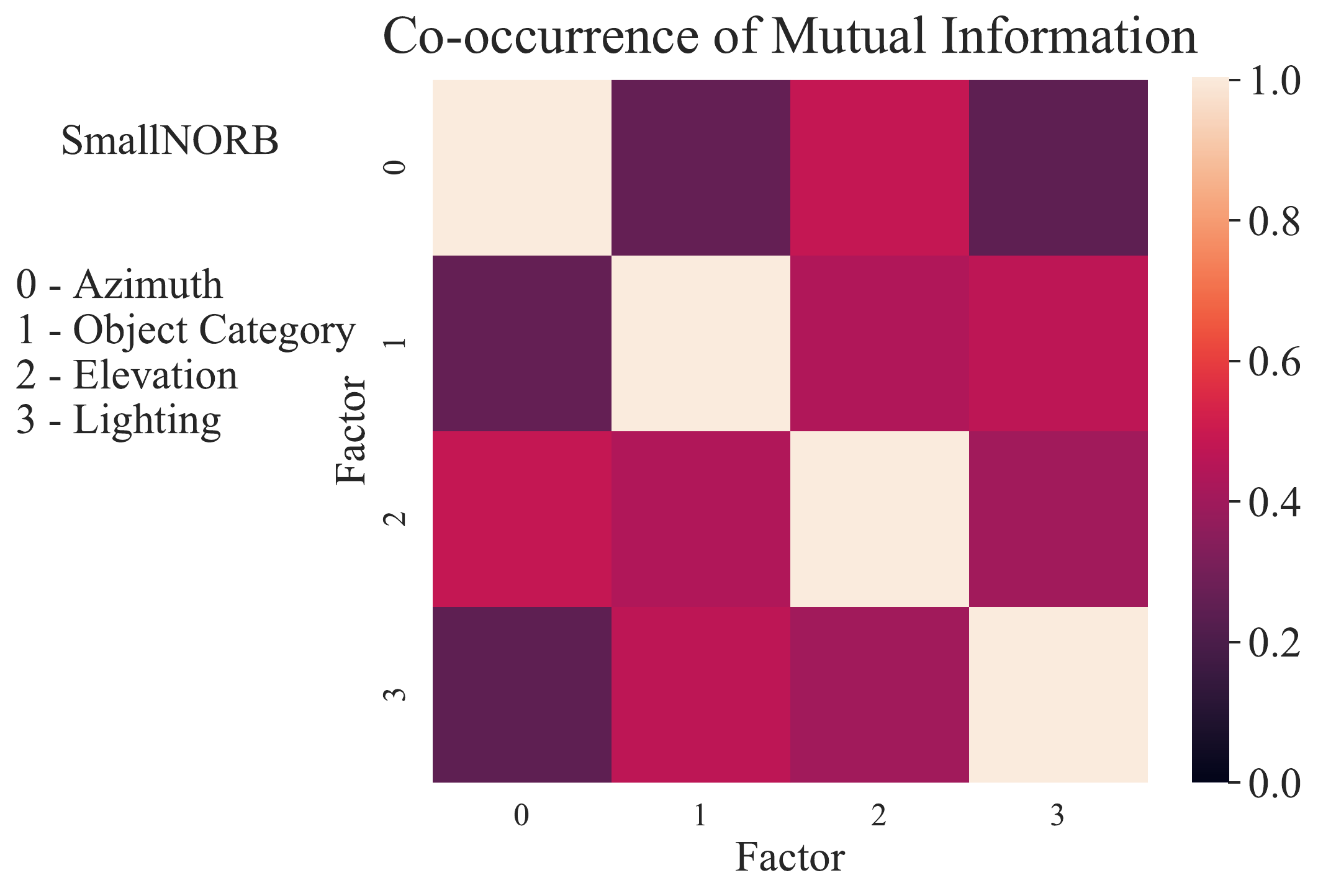}
  \caption{}
  \label{fig:cooccurence_smallnorb}
\end{subfigure}
\begin{subfigure}{.22\textwidth}
  \centering
  \revfig{\includegraphics[width=\linewidth]{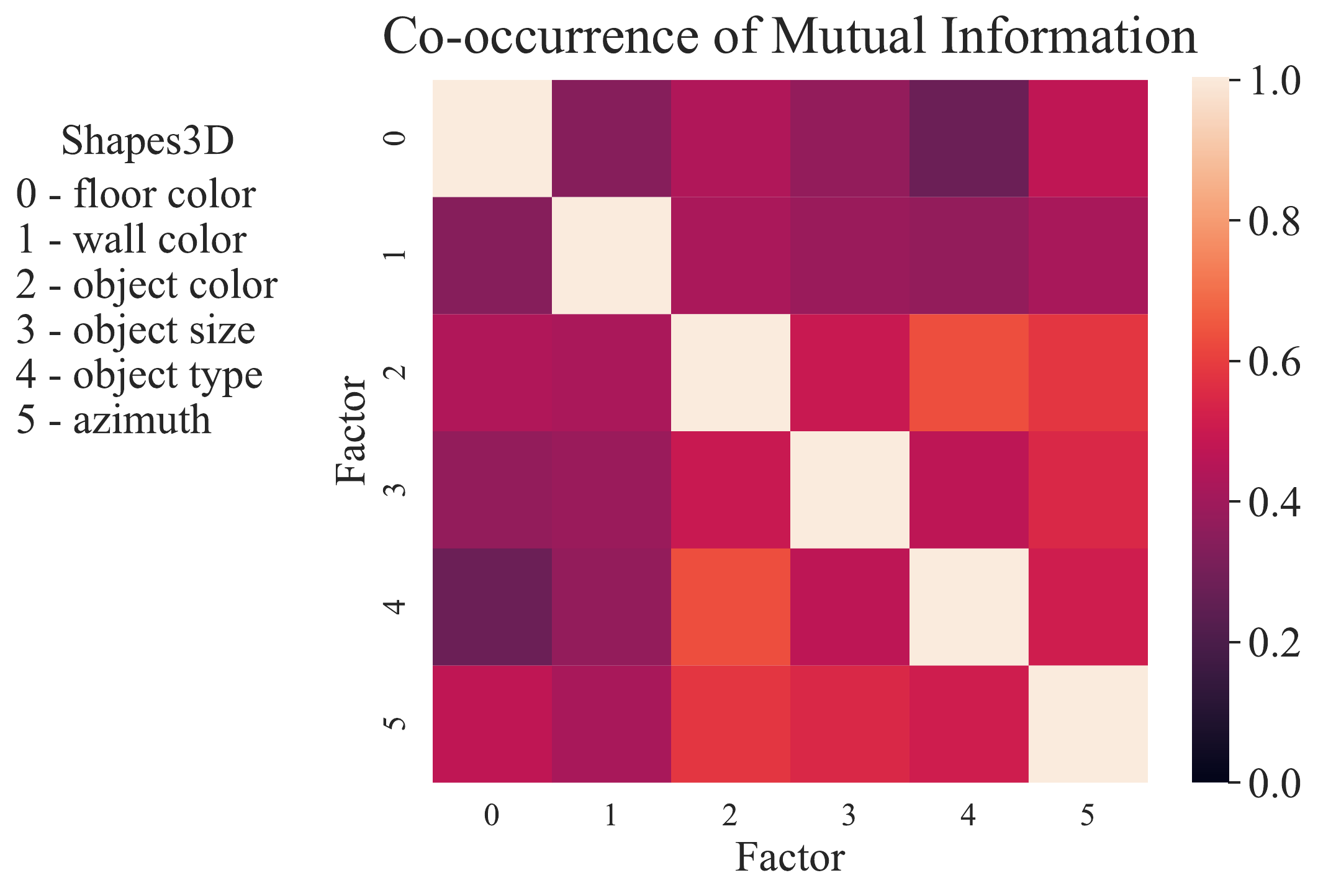}}
  \caption{}
  \label{fig:cooccurence_shapes3d}
\end{subfigure}
\begin{subfigure}{.22\textwidth}
  \revfig{\includegraphics[width=\linewidth]{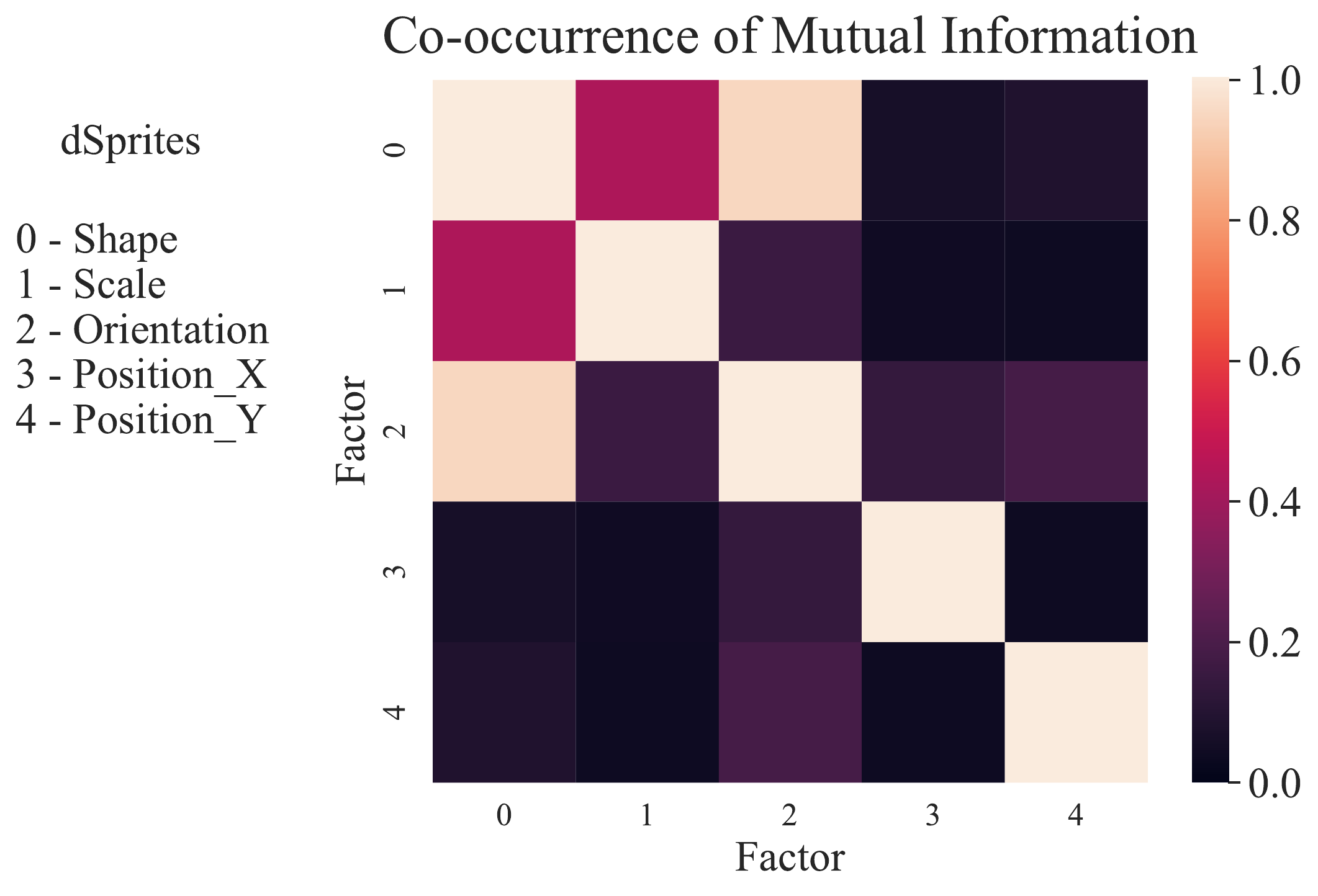}}
  \caption{}
  \label{fig:cooccurence_dsprites_full}
\end{subfigure}
\caption{The co-occurrence of factors in the mutual information relationship among BYOL representations on Cars3D(a), SmallNORB(b) , Shapes3D(c), \revtext{and dSprites including all factors(d).}}
\label{fig:co_full}
\end{figure}

\begin{figure}
    \centering
    \includegraphics[width=.68\linewidth]{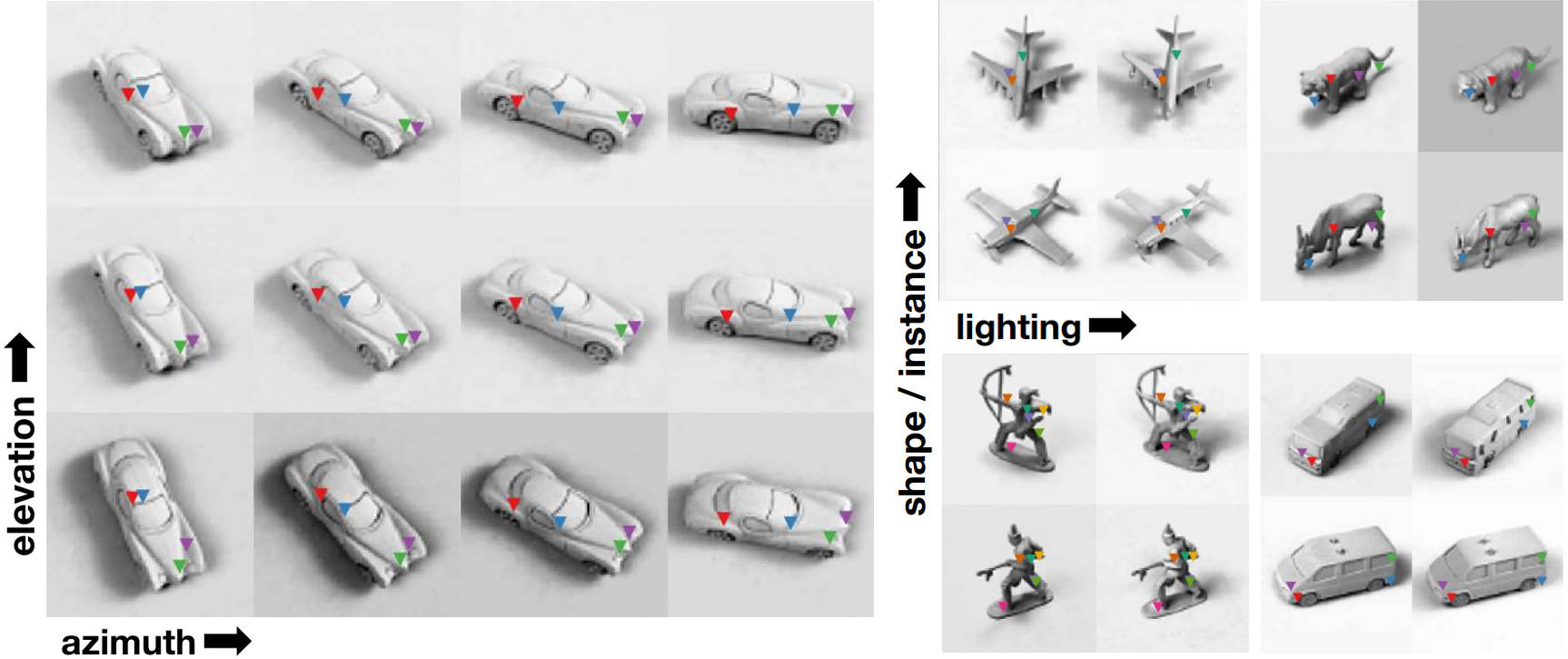}
    \caption{Some samples from SmallNORB dataset. The variance is controlled by the factor indicated on the axis. The image is from ~\citet{jakab2018unsupervised}.}
    \label{fig:smallnorb_sample}
\end{figure}

\subsection{Importance Distribution by DCI}
In the main text, we concisely talked about the potential variables introduced by the learnable model under some metrics. Here we show an example for the widely used DCI Disentanglement metric. We follow DisLib to use Gradient Boosting Tree to estimate the importance matrix between each factor and each latent dimension. All parameters are set the same as its default protocol. The visualization is shown in Figure~\ref{fig:importance_distri}. Compared with the mutual information distribution shown in Figure~\ref{fig:separation}, the importance distribution is much more sparse. 
This is because the construction of the GBT regressor encourage the sparsity of the output importance matrix. Such a sparsity can lead to the misunderstanding that the correlation between factors and latent dimensions is also sparse which is not true. By using pure measurement without involving additional adaptive models, such a problem will not be raised in the proposed MED metric.
\subsection{Mutual Information Heatmaps}
\label{sec:MI_heatmaps}
We compute MI between each latent dimension and each data factor and visualize them by heatmaps. The heatmaps offer us an intuitive picture of the learned representation space.  For completeness, we show the MI heatmaps on SmallNORB, Shapes3D, and \revtext{dSprites with all factors} in Figure~\ref{fig:mi_smallnorb}, Figure~\ref{fig:mi_shapes3d} and \revtext{Figure~\ref{fig:mi_dsprites_full}} respectively. We can see that the disentangled pattern described in the main text still emerges. There is a group of columns brighter than others in each row, and these groups do not overlap for most rows. However, we find that some latent dimensions emphasize more than one factor. We provide a more detailed analysis from the perspective of factor co-occurrence on this phenomenon in~\ref{sec:co} below. 

\subsection{Co-occurrence of Factors}
\label{sec:co}
To understand to what extent one dimension of the learned representation would respond to more than one factor, we make the co-occurrence of mutual information to factors on more datasets here. The visualizations are shown in Figure~\ref{fig:co_full} for the results on Cars3D, SmallNORB, Shapes3D, and dSprites respectively. Moreover, we now analyze the definition of data factors of these datasets to discuss whether they are defined to be fully independent or not.
\begin{wrapfigure}{R}{0.55\textwidth}
  \begin{center}
    \includegraphics[width=0.55\textwidth]{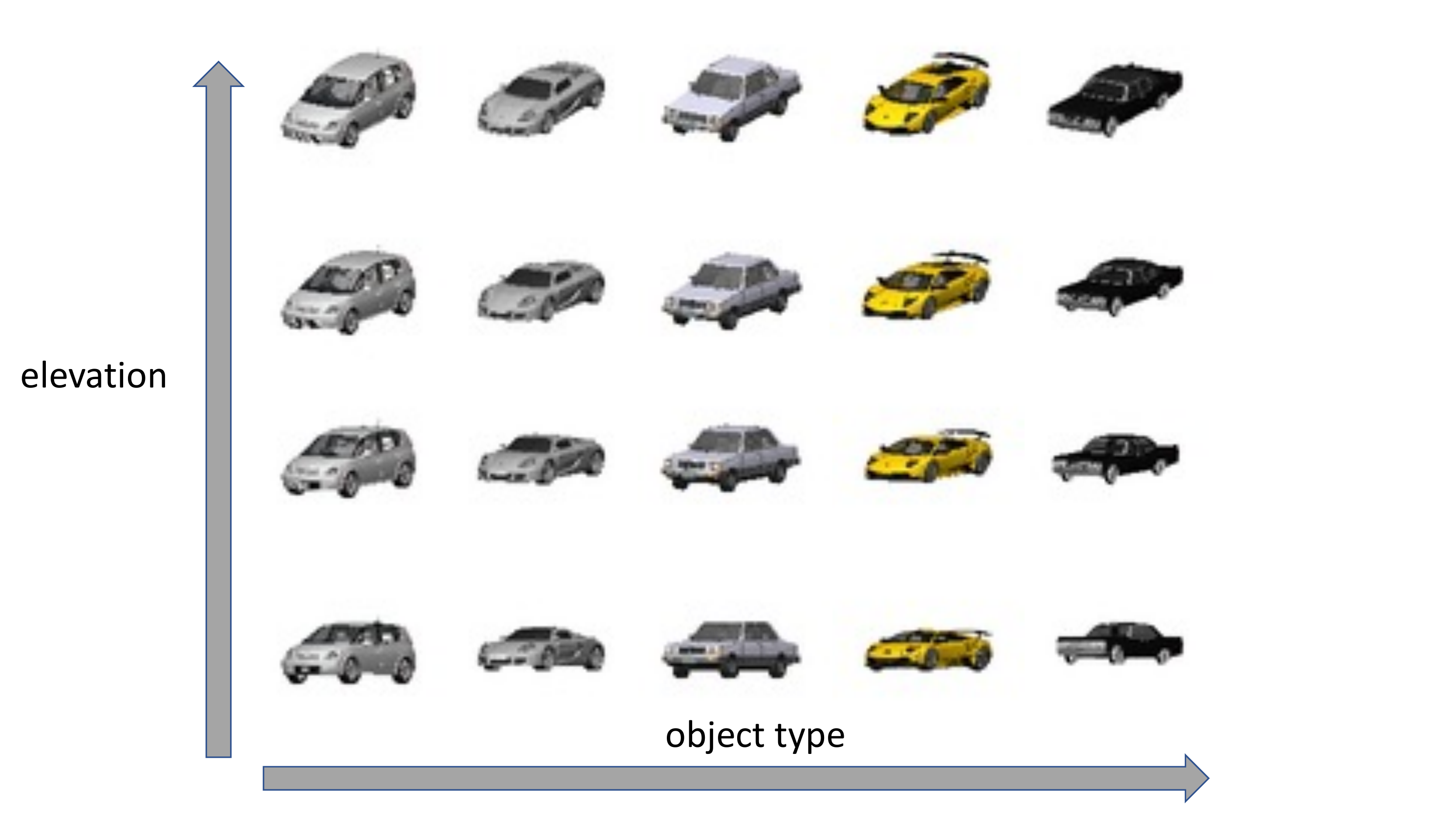}
  \end{center}
  \caption{Samples from Cars3D. The variables of the object type and the elevation are controlled. It shows that the two factors are not independent.}
  \label{fig:cars3d_sample}
\end{wrapfigure}
\paragraph{SmallNORB} Though most non-diagonal entries have very low co-occurrence of mutual information, two pairs of factors show slightly higher co-occurrence. They are ``azimuth-elevation'' and ``instance category-lighting''. 
After investigating the dataset, we find the two pairs of factors are not fully independent. 
Figure~\ref{fig:smallnorb_sample} show some samples with corresponding factors manipulated. We could see that the elevation and azimuth are not fully independent. And the correlation between the instance category and the lighting factor is even more obvious because the lighting condition is sensibly related to the shadow around the object, whose distribution and shape are highly determined by the instance category. 

\paragraph{Cars3D} Only one pair of factors show some co-occurrence, i.e. ``elevation-object type''. We randomly selected samples from Cars3D by different object types and elevations, as shown in Figure~\ref{fig:cars3d_sample}. It shows that with the same value of elevation, samples of different object types have different visual elevations. So these two factors are not fully independent. This explains the slightly higher co-occurrence of mutual information between this pair of factors.

\paragraph{Shapes3D} The result shows relatively bad disentanglement. To be precise, some factor pairs show low mutual information co-occurrence as expected, such as the color factors of floor, wall, and object and the pair of ``object color - azimuth''. But the MI co-occurrence of ``wall color - object size'' and ``object color - object size;'' are higher than we expected as we did not recognize their high dependence. This result might relate to our model's relatively poor performance on Shapes3D. 

\revtext{\paragraph{dSprites} We omit the factor \emph{orientation} in the visualization at main text for clarity. Here we show the results including all factors. For dSprites dataset, the value of \emph{orientation} is highly correlated with \emph{shape}. This is because that different \emph{shape} has different symmetry properties for rotation.  As a result, dimensions that encode information of \emph{shape} tend to capture information of \emph{orientation} simultaneously. This explains that, in Figure~\ref{fig:mi_dsprites_full}, some dimensions have two brighter entries at ``shape'' and ``orientation'' rows. Thus, it is not surprising that the ``shape - orientation'' cell is brighter than other off-diagonal elements in Figure~\ref{fig:cooccurence_dsprites_full}. 
}

\subsection{Manipulating Factors}
\label{sec:appendix_mani_factors}
\begin{figure*}[ht]
    \centering
    \includegraphics[width=0.9\textwidth]{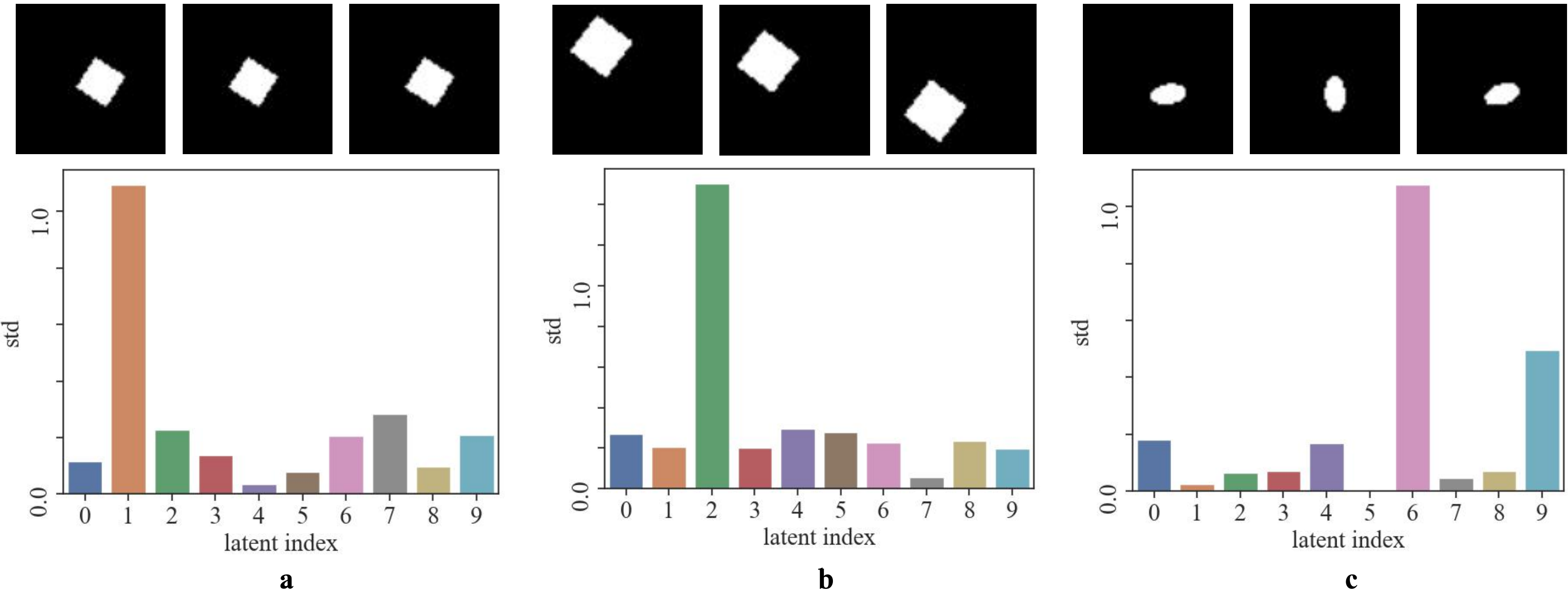}
    \caption{Representation variation when manipulating one factor only in the dimension-reduced version. In (a) and (b), \textit{position\_x} and \textit{position\_y} are manipulated respectively and only cause one dimension significantly variate. While, in (c), when manipulating the ill-defined factor \textit{orientation}, two dimensions variate.}
    \label{fig:factor_pca_std}
\end{figure*}

In the main paper, we studied the influence on the generated representation by manipulating the factors, where the representation is reduced by selecting dimensions as in calculating Top-k MED. Here, we do the qualitative study of the influence on representation by manipulating factors in another way but still on dSprites. To make the original high-dimensional representation space more compact, we use the unsupervised dimension reduction by PCA instead, which is more general when the data factor pattern is unknown.
Here, we reduce the representation dimension by PCA to 10. Note that since the PCA operation mixes the original latent space with a linear combination, it might destroy the existing disentanglement property in the high dimensional space, or enhance the disentanglement if the original high dimensional space is a linear combination of the ground truth factors. But such influence is usually considered secondary to the disentanglement learned by a model. No matter which case, if the dimension-reduced representation shows disentangled properties, the original space at least captures linearly transformed ground truth factors, and the dimension reduction techniques such as PCA can make the representation more compact in a qualitative study.

Figure~\ref{fig:factor_pca_std} shows the result of representation vector variation when changing only one factor at once. Given three images with only one factor's value being different, we generate the 10-dim representation vectors from them. Then, we compute the variance across the three vectors, leading to 10 scalars. The larger the variance is, the more that dimension responds to the factor change. Figure~\ref{fig:factor_pca_std}(a) and (b) show how reduced representation vector changes when manipulating \textit{position\_x} and \textit{position\_y} factor respectively. 
It shows good disentanglement that only one representation dimension has high variation. However, in Figure~\ref{fig:factor_pca_std}(c)  we show a failure mode of the ill-defined factor \textit{orientation} that change of factor causes both the 6th and the 9th dimensions of reduced representation to have large variations. 
From the results, we observe that manipulating one well-defined independent factor causes evident variance in only one dimension. And it shows that we could make the learned representation vector more compact by unsupervised dimension reduction.

\newcommand{\utextbf}[1]{\underline{\textbf{#1}}}
\begin{table*}[!htp]
\centering
\small
\renewcommand{\arraystretch}{1.1}
\caption{Evaluation results on multiple datasets with different disentanglement metrics. The best results of all models are \textbf{bolded}. The best results on low-dimensional space are \underline{underlined}.}
\begin{tabular}{p{0.2cm}|l|r|r|r|r|r|r} 
     \toprule
    & Model & BetaVAE & FactorVAE & MIG & SAP & DCI & \newmet{}\\ 
     \midrule 
     % VAE & 0.632 & - & - & - \\
     \multirow{11}{*}{\begin{turn}{-270}\textbf{dSprites}\end{turn}} & $\beta$-VAE & 82.3 (7.6) & 65.8 (9.2) & 26.3 (11.0) & 5.2 (2.7) & 39.3 (13.2) & 32.6 (10.0)\\
     & $\beta$-TCVAE  & 86.7 (2.4) & 76.6 (7.8) & 23.8 (6.8) & 6.9 (0.9) & 36.3 (7.1) & 31.8 (7.4)\\
     & FactorVAE & 84.9 (2.8) & 75.3 (7.4) & 18.4 (9.0) & 6.8 (0.8) & 28.8 (10.6) & 32.5 (10.1)\\
     & DIP-VAE-I  & 82.7 (3.3) & 59.1 (4.8) & 9.6 (5.1) & 5.2 (2.6) & 14.4 (4.6) & 18.8 (5.6)\\
     & DIP-VAE-II & 81.5 (4.9) & 58.6 (7.6) & 7.4 (3.4) & 3.6 (2.2) & 12.3 (5.2) & 14.7 (5.5)\\
     & AnnealedVAE & 86.5 (0.1) & 60.1 (0.0) & \underline{\textbf{35.2 (1.3)}} & \underline{7.6 (0.5)} & 37.9 (2.1) & 35.8 (0.8)\\
     & Ada-GVAE & 88.0 (2.7) & 73.1 (3.9) & 17.3 (4.7) & 6.6 (2.0) & 32.3 (4.6) & --\\
     & SlowVAE  & 87.0 (5.1) & 75.2 (11.1) & 28.3 (11.5) & 4.4 (2.0) & \underline{47.7 (8.5)} & --\\
      & EBM & 82.3 (2.0) & 65.7 (12.5) & 1.7 (0.5) & 3.0 (1.2) & 19.1 (1.8) & 6.8 (4.0)\\
     & InfoGAN-CR & 85.5 (1.0) & \underline{88.0 (1.0)} & 19.8 (3.2)& 6.0 (1.0)& 14.0 (5.2) & --\\
     & \CC{100} BYOL & \CC{100} \textbf{93.2 (0.4)} & \CC{100} \textbf{91.6 (0.8)} & \CC{100} 29.3 (0.4)  & \CC{100} \textbf{8.0 (0.4)}  & \CC{100} \textbf{66.9 (0.2)} & \CC{100} 31.3 (0.4) \\
     & \CC{100} \revtext{BYOL (top-2)}& \CC{100} \underline{90.0 (0.9)} & \CC{100} 79.2 (2.3) & \CC{100} 3.3 (0.9) & \CC{100}  0.8 (0.2) & \CC{100} 45.0(0.1) & \CC{100} \underline{\textbf{53.7 (0.7)}}\\
     \midrule
     \multirow{11}{*}{\begin{turn}{-270}\textbf{Cars3D}\end{turn}}  & $\beta$-VAE & \utextbf{100.0 (0.0)} & 89.3 (1.2) & 11.7 (1.1) & 1.4 (0.9)  & 38.7 (4.6) & 29.0 (2.2)\\
     & $\beta$-TCVAE  & \utextbf{100.0 (0.0)} & 92.2 (2.7) & \utextbf{15.5 (2.9)} & 1.7 (0.3) & 42.7 (3.5) & \utextbf{33.0 (3.8)}\\
     & FactorVAE & \utextbf{100.0 (0.0)} & 91.7 (4.1) & 10.6 (2.2) & \utextbf{2.0 (0.5)} & 29.0 (6.7)& 29.1 (3.0)\\
     & DIP-VAE-I  & \utextbf{100.0 (0.0)} & 90.5 (5.0) & 5.9 (2.8) & 1.9 (1.4) & 22.6 (5.6) & 19.4 (3.3)\\
     & DIP-VAE-II & \utextbf{100.0 (0.0)} & 85.0 (6.1) & 5.1 (2.7) & 1.3 (0.8) & 20.8 (5.4) & 16.7 (4.1)\\ 
     & AnnealedVAE & \utextbf{100.0 (0.0)} & 85.0 (4.3) & 7.6 (1.0) & 1.5 (0.5) & 18.5 (4.3) & 15.5 (2.5)\\
     & SlowVAE  & \utextbf{100.0 (0.0)} & 90.4 (0.5) & 15.4 (2.2) & 1.6 (0.5) & \underline{48.0 (2.4)} & --\\
     & \CC{100} BYOL & \CC{100} \utextbf{100.0 (0.0)} & \CC{100} \textbf{95.8 (1.2)} & \CC{100} 7.6 (0.9) & \CC{100} 1.8 (0.7) & \CC{100} \textbf{48.5 (2.3)} & \CC{100} 9.7 (0.5)\\
     & \CC{100} \revtext{BYOL (top-3)} & \CC{100} \utextbf{100.0 (0.0)} & \CC{100} \underline{95.2 (0.8)} & \CC{100} 3.8 (0.5) & \CC{100} 1.1 (0.8) & \CC{100} 15.8 (3.6) & \CC{100} 31.8 (1.3)\\
     \midrule 
     \multirow{10}{*}{\begin{turn}{-270}\textbf{SmallNORB}\end{turn}}  & $\beta$-VAE & 84.1 (2.7) & 60.1 (2.4) & 25.0 (1.1) & 11.4 (1.1) & 32.6 (0.6) & 24.4 (0.7)\\
     & $\beta$-TCVAE  & 84.5 (2.7) & 60.3 (2.3) & 25.4 (0.9) & 11.7 (1.1) & \underline{35.2 (0.7)} & 25.0 (0.9)\\
     & FactorVAE & 80.8 (3.8) & 62.5 (3.6) & 23.9 (2.0) & 10.2 (0.9) & 33.4 (1.1) & \utextbf{25.9 (1.2)}\\
     & DIP-VAE-I  & 84.2 (3.2) & \underline{69.8 (4.6)} & 24.3 (2.7) & 10.2 (1.4) & 30.0 (2.1) & 24.5 (2.1)\\
     & DIP-VAE-II & 85.2 (1.3) & 58.4 (2.1) & \utextbf{25.5 (1.5)} & \utextbf{14.4 (0.4)} & 32.3 (0.7) & 24.4 (0.7)\\
     & AnnealedVAE & 60.8 (6.2) & 50.0 (9.9) & 9.1 (2.2) & 6.8 (0.8) & 15.7 (6.4) & 5.5 (3.7)\\
      & SlowVAE  & 78.2 (3.8) & 47.0 (2.9) & 23.8 (1.8) & 7.8 (1.1) & 28.7 (0.7) & 21.8 (1.3)\\
      & EBM & 79.0 (4.4) & 57.9 (3.5) & 1.7 (0.5) & 1.9 (0.1) & 13.9 (2.2) & 2.3 (1.7) \\
      & \CC{100}  BYOL & \CC{100} \textbf{97.0 (0.8)} & \CC{100} \textbf{81.0 (0.5)} & \CC{100} 3.3 (0.9) & \CC{100} 2.2 (0.3) & \CC{100} \textbf{51.0 (1.0)} & \CC{100} 7.7 (0.2)\\
      & \CC{100}  \revtext{BYOL (top-2)} & \CC{100} \underline{86.7 (0.4)} & \CC{100} 65.6 (3.7) & \CC{100} 3.3 (1.4) & \CC{100} 1.5 (0.2) & \CC{100} 13.6 (0.3) & \CC{100} 25.7 (0.3)\\
      \midrule 
      \multirow{11}{*}{\begin{turn}{-270}\textbf{Shapes3D}\end{turn}} & $\beta$-VAE & 100.0 (0.0) & 92.4 (4.5) & 37.8 (16.0) & 11.3 (3.2) & 77.3 (3.2) & 52.4 (9.4) \\
     & $\beta$-TCVAE  & 100.0 (0.0) & 90.5 (5.5) & 46.4 (15.4) & 12.4  (6.1) & 78.4 (5.2) &  53.2 (4.9) \\
     & FactorVAE & 98.1 (3.2) & 90.6 (6.4) & 48.2 (15.2) & 11.1 (4.3) & 71.8 (8.6) & 55.9 (8.0)\\
     & DIP-VAE-I  & 98.3 (3.3) & 84.2 (11.3) & 20.1 (8.4) & 6.0 (1.1) & 69.0 (3.6) & 43.5 (3.7)\\
     & DIP-VAE-II & 99.6 (0.04)  & 94.9 (4.1) & 22.1 (3.54) & 8.2 (1.8) & 49.8 (10.6) & 52.6 (5.2)\\
     & AnneledVAE & 95.1 (4.4) & 88.8 (4.6) & 46.2(5.4) & 8.5(1.6) & 56.2(4.7) & \utextbf{56.1 (1.5)}\\ 
     & Ada-ML-VAE & \utextbf{100.0} & \utextbf{100.0} & 50.9 & 12.7 & 94.0 & --\\ 
     & Ada-GVAE & \utextbf{100.0} & \utextbf{100.0} & 56.2 & \utextbf{15.3} & \utextbf{94.6} & --\\
      & SlowVAE  & \textbf{100.0 (0.1)} & 97.3 (4.0) & \utextbf{64.4 (8.4)} & 5.8 (0.9) & 82.6 (4.4) & --\\
      & EBM & 75.9 (11.2) & 53.2 (8.7) & 5.2 (2.2) & 2.8 (1.1) & 21.8 (11.0) & 2.1 (2.6) \\
      & \CC{100}  BYOL & \CC{100} 91.5 (3.9) & \CC{100} 82.5 (2.4) & \CC{100} 5.2 (1.7) & \CC{100} 2.8 (0.3) & \CC{100} 53.1 (1.5) & \CC{100} 6.0 (0.5)\\
      & \CC{100}  \revtext{BYOL (top-2)} & \CC{100} 95.5 (1.1) & \CC{100} 83.0 (3.6) & \CC{100} 2.8 (0.8) & \CC{100} 1.4 (0.5) & \CC{100} 27.2 (3.8) & \CC{100} 19.7 (1.3)\\
      \midrule
      \multirow{8}{*}{\begin{turn}{-270}\textbf{CelebA}\end{turn}} & VAE & 21.5 (3.2) & 6.1 (3.8) & 0.8 (0.1) & 0.9 (0.2)& 11.2 (2.3) & 3.8 (0.2)\\
     & $\beta$-VAE                      & 19.1 (1.9) & 5.8 (1.8)         & 0.1 (0.1) & 0.6 (0.2) & 8.7 (1.9) & 3.3 (0.1)\\
     & $\beta$-TCVAE                    & 19.9 (2.3) & 9.8  (2.4)       & 0.6 (0.2)& 1.2 (0.3) & 3.5 (1.1)& 4.7 (0.1)\\
     & FactorVAE                        & 25.3 (3.0) & \utextbf{12.0 (2.1)} & 0.4 (0.1)& 0.6 (0.2) & 7.1 (0.7) & 0.6 (0.6)\\
     & DIP-VAE-I                        & 21.0 (1.9) & 9.3 (1.1)         & 0.2 (0.1)& 0.9 (0.3)& 13.8 (2.2) & 3.6 (0.2)\\
     & InfoGAN-CR & 16.8 & 11.3 & \underline{1.6} & \underline{2.8} & \underline{22.0} & -- \\
     & \CC{100}  BYOL & \CC{100} \textbf{35.7 (2.1)} & \CC{100} 11.5 (1.1) & \CC{100} \textbf{2.6 (0.7)}  & \CC{100} \textbf{8.2 (0.9)} & \CC{100} \textbf{41.0 (1.3)} & \CC{100} 4.8 (0.4) \\
     & \CC{100}  \revtext{BYOL (top-3)} & \CC{100} \underline{25.8 (0.8)} & \CC{100} 9.7 (0.5) & \CC{100} 0.7 (0.2)  & \CC{100} 0.2(0.02) & \CC{100} 15.8(0.3) & \CC{100} \utextbf{6.8 (0.7)} \\
      \bottomrule
\end{tabular}
\label{table:full_results}
\end{table*}

\section{More Quantitative Results}
\label{sec:appendix_more_quan}
\revtext{\subsection{Full Disentanglement Benchmark}}
\label{sec:appendix_more_benchmark}
In the main text, we evaluate the disentanglement under our proposed MED metric \revtext{for full space and top-k subspace} on multiple datasets. In this section, to provide a more complete understanding of the disentanglement property of contrastive learning without negatives, we report the disentanglement scores with other metrics, including FactorVAE score, BetaVAE score, MIG, SAP, and DCI Disentanglement. 

For the results of VAE-based methods, as the large-scale benchmark of \citet{locatello2019challenging} provides the original logs on dSprites, Cars3D, and SmallNORB datasets, we simply report the performance of the best configuration. 
The original logs on Shapes3D are not available, so we train and evaluate on Shapes3D by ourselves for the MED scores. 
For scores under other metrics, we report the median disentanglement scores. Some results are from~\citet{locatello2020weakly} but the std error is not available. The median performance of SlowVAE is from its original paper~\citep{slowvae}. For the results of CelebA, the result of InfoGAN-CR is from its officially released checkpoint without availability to the std error. For other methods, we report the mean value of our trained weights over three random seeds as default. Because the evaluation of DCI is extremely time-consuming, around 14 hours for a 1000-d model, we only take BYOL as an example here for negative-free contrastive learning methods.  
All results are combined and shown in Table~\ref{table:full_results}. \revtext{We also include traditional metrics results on their selected top-k subspace for contrastive learning methods.}

Aligned with the analysis we provide in the main text, the results show significant disagreement among the existing metrics. To be precise, for those metrics (BetaVAE score, FactorVAE score, DCI Disentanglement) using a learnable model such as a regressor or a classifier, the high-dimensional BYOL model achieves a significant advantage. However, for the metrics relying on only one or two dimensions to reveal the connection between a latent dimension and a factor (MIG and SAP), BYOL's performance is not that impressive anymore.

Finally, the result on CelebA shows the great robustness of BYOL's learned representations to be disentangled on real-world datasets. Yet, the large gap between the score \revtext{of CelebA and} those on synthetic datasets emphasizes the difficulty of learning disentangled factors on real-world images. It is hard to empirically study whether it is the high dimension that gives BYOL advantages on some metrics because the nature of BYOL makes it hard to be trained with a small latent dimension to make a comparison. 

\revtext{In the last rows for each dataset, we show previous disentanglement metrics scores on the top-k subspaces of BYOL. We find that BYOL (top-k) has performed better than or on par with unsupervised methods with low latent dimensions. Note that Ada-GVAE, SlowVAE, and EBM are (weakly) supervised methods. The SOTA performance justifies our claim of the well-disentangled subspaces. MIG and SAP require a large gap between the two most vital representation dimension-factor responses, measured by mutual information and classification accuracy. But out of 1000 dimensions, we choose the k most relative ones. Hence, the gap ought to be minimal. Consequently, BYOL (top-k) receives lower MIG and SAP scores. Except for dSprites and CelebA, BYOL (top-k) have lower DCI scores. The primary cause of this is the massive information loss during the selection of the top-k dimensions, which completely obliterates the original encoding pattern. As a result, the gradient boosting tree (GBT) developed throughout the DCI process tends to ``borrow" information from latent dimensions that emphasize additional relative factors to classify some complex factors. The ``information-lending" dimensions then become less disentangled since they are also important for predicting those complex factors.  Taking the Cars3D dataset as an example, we set $k=3$ for top-k MED. To predict the factor \emph{object type}, the GBT needs to classify 183 types of cars from only 9 dimensions with only 3 emphasizing \emph{object type}. This forces GBT to utilize information from \emph{elevation} since these two attributes are correlated (see Appendix~\ref{sec:co}). However, we keep all they have learned from training for other low-dimensional methods. Therefore, the information loss problem does not apply to them.}

% \revtext{In summary, we obtain the following conclusions regarding metrics and top-k subspace with the complete benchmarks. \textbf{(1)} The scores of the whole space of BYOL empirically verify the disagreement of previous metrics. \textbf{(2)} Traditional metric scores on the top-k subspace support the well-disentangle property of the subspace.}

\section{Ablation Study about the Normalization Choice}
\label{sec:appendix_ablation}
In this section, we put ablation studies here about the choice of normalization layers in the model architecture.

\begin{table*}[!htp]
\centering
\caption{Results of using different normalization strategies on dSprites.}
\begin{tabular}{ c|p{1.5cm}<{\centering}|p{1.5cm}<{\centering}|p{1.5cm}<{\centering}|p{1.5cm}<{\centering}|p{1.5cm}<{\centering}  } 
     \toprule
      \textbf{normalization} & w/o norm &  BN & GN & LN & IN\\ 
     \midrule
     \textbf{\newmet{}} & 23.8 (0.6) & 29.4 (0.5) & 31.3 (0.4) & 31.3 (0.8) & 0.0 (0.0) \\
     \bottomrule
\end{tabular}
\label{table:norm_dsprites}
\end{table*}

We experiment with five types of normalization layers in the encoder network on the dSprites dataset. The results are shown in Table~\ref{table:norm_dsprites}.  For the group normalization, we set the group number to 4. 
On the dSprites dataset, we find the commonly used BN decreases the disentanglement performance. By keeping the batch norm in the projector and the predictor, removing the batch norm in the encoder will not cause the model to collapse, which agrees with the observation in previous works~\citep{richemond2020byol}. On the contrary, replacing batch norm in the encoder with group norm or layer norm will increase the representation disentanglement while achieving similar accuracy in downstream factor prediction. We notice that a similar phenomenon has been discovered before in supervised representation disentanglement. For example, \citet{bau2017network} discovered that a network trained with batch normalization layers has less interpretable (disentangled) neurons. 
On the other hand, the instance norm~\citep{ulyanov2017improved} completely breaks the contrastive learning process. We still do not fully understand this behavior, but we hypothesize that it may be caused by the shared batch statistics that make it hard for a feature to be aligned with the ground truth factor.

\section{Disagreement of Existing Metrics}
\label{sec:disagreement_appendix}
\begin{figure}
    \begin{center}
        \revfig{\includegraphics[width=.6\textwidth]{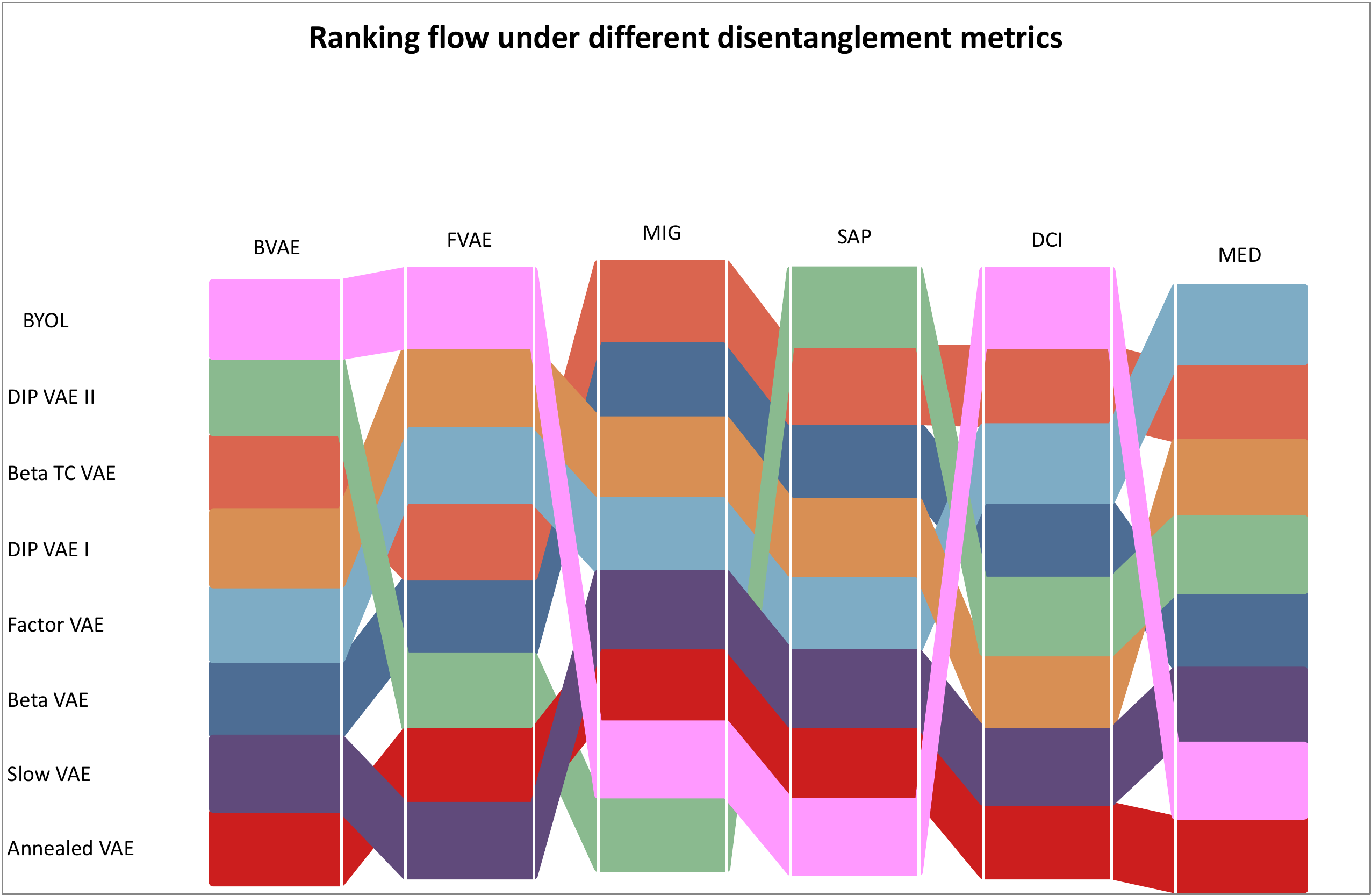}}
    \end{center}
    \caption{\revtext{The flow chart of rankings of disentanglements scores. From top to down, the ranking decreases, indicating high to low disentanglement properties. If all metrics agree perfectly, there should be NO switch of relative ranking among these methods with different metrics. However, we notice obvious ranking variance, especially regarding the ranking of BYOL. With the analyses of the bias of existing metrics when the representation dimension varies, this figure emphasizes the necessity of proposing a new disentanglement metric that is fair to compare models of different dimensions.}}
    \label{fig:flow_rank}
\end{figure}

Without a uniformly recognized definition of ``disentanglement'', the validity of existing metrics is backed up with their shared expectation of ``disentanglement'' and consistency of experimental results, as suggested in DisLib~\cite{locatello2019challenging}.  However, the results in Figure~\ref{fig:disagreement} break the belief. The disagreement arises when the high-dimensional representation model joins the comparison. 
We provide an in-depth view of the disagreement by ranking different methods. The results are shown in Figure~\ref{fig:flow_rank}. If all metrics agree perfectly, there should be no relative ranking switch. However, the frequent switch, especially with BYOL being taken into comparison, strongly suggests their disagreement. Results in Table~\ref{table:full_results} further show that different metrics may make completely different rankings with large value gaps. We also provide  logistic analysis in Appendix~\ref{sec:superority_of_med} and quantitative results in Appendix~\ref{sec:appendix_more_quan}. Given the fact that self-supervised representation learning always requires a high dimension to train, MED is necessary to extend the study of disentangled representation learning to complicated real-world datasets and self-supervised representation learning methods.

\section{\revtext{The Superiority of MED}}
\label{sec:superority_of_med}
\revtext{Here, we provide both experimental observations and theoretical analysis in synthetic scenarios to show the superiority of MED: in what cases other metrics outputs meaningless or even opposite results but MED can still perform a meaningful evaluation.
}

\subsection{\revtext{Experimental Observations}}
As we discuss throughout the paper, the five existing metrics we investigate, i.e. BetaVAE score, FactorVAE score, SAP, MIG, and DCI Disentanglement, are unfair to models of different dimensions. So we extend the results mentioned in Section~\ref{sec:med} to show the superiority of MED by confirming its stability under different dimensions. We conduct a sanity check on randomly initialized models. For a good disentanglement metric, we do not expect a high score whatever the representation dimension is. The following results reveal the malfunction of these metrics.
\begin{itemize}
    \item \textbf{MED vs MIG/SAP}: First of all, we emphasize that SAP/MIG is designed not just for \emph{disentanglement} but also \emph{completeness}. They encourage a strict one-to-one mapping between latent dimensions and factors. So they tend to reward models whose dimension is close to the factor number. On the popular synthetic datasets with only a limited number of factors, low-dimensional models have an unfair advantage with MIG/SAP metrics. We compare a well-trained 1000-d BYOL model and a 10-d randomly initialized encoder in Table~\ref{tab:wrong_table_2}. The results show that the 10-d randomly initialized model can achieve higher scores than the 1000-d BYOL model by SAP and MIG. But MED keeps a reasonable comparison that the trained high-dimensional model can still enjoy higher scores than low-dimensional random weights.
    \begin{table}[hb]
    \caption{\revtext{The mean scores of different metrics on Shapes3D.}}
        \centering
        \revfig{
        \begin{tabular}{l|c|c|c}
        \toprule
            & MIG & SAP & MED\\
            \midrule
            10-d randomly initialized & \textbf{6.7} & \textbf{3.0} & 2.9\\
            \hline
            1000-d BYOL & 5.2 & 2.8 & \textbf{6.0}\\
            \bottomrule
        \end{tabular}}
        \label{tab:wrong_table_2}
    \end{table}
    \item \textbf{MED vs BetaVAE/FactorVAE score}: BetaVAE and FactorVAE prefer high-dimensional models as they have more parameters to trick the learnable classifier. In Table~\ref{tab:wrong_table_1}, we find even randomly initialized high-dimensional (1000-d) model can achieve higher BetaVAE/FactorVAE score than a well-trained 10-d model (DIP-VAE-II). In contrast, MED provides results in line with our expectations.
    \begin{table}[hb]
    \caption{\revtext{The mean scores of different metrics on dSprites.}}
        \centering
        \revfig{
        \begin{tabular}{l|c|c|c|c}
        \toprule
            & BetaVAE score & FactorVAE score & DCI & MED\\
            \midrule
            10-d DIP-VAE-II & 81.5 & 58.6 & 12.3 & \textbf{14.7}\\
            \hline
            1000-d randomly initialized & \textbf{82.7} & \textbf{61.4} & \textbf{19.8} & 3.8\\
            % \hline 
            % 1000-d BYOL & 93.2 & 91.6 & 66.9 & 31.3\\
            \bottomrule
        \end{tabular}}
        \label{tab:wrong_table_1}
    \end{table}
    \item \textbf{MED vs DCI}: Similar to BetaVAE/FactorVAE scores, DCI also overestimate high-dimensional models as they have a higher chance to get lottery dimensions, especially with the sparsity encouraged by GBT. Also, another huge drawback of DCI is the computation efficiency: for a 1000-d model, DCI takes more than 14 hours for evaluation while MED only takes less than 20 seconds. The computation of MED is more than 2000 times faster than DCI.
\end{itemize}

\revtext{Through the experimental observations, we find the abnormals from existing metrics when comparing models of different dimensions. Compared to them, the results of MED are consistent with human knowledge.}

\subsection{\revtext{Theoretical Analysis}}
\revtext{In this part, we construct synthetic scenarios to investigate the validity of MED and existing metrics. We will show that in such situations, existing metrics will fail to reflect the real disentanglement quality but MED can show superiority to maintain reasonable evaluations. For the following cases, we assume there are two data factors $F_0, F_1$ on the data. We note their value as $v_0, v_1 \in \{0,1\}$. And they are sampled independently and uniformly, i.e., $p(v_j=1) = p(v_j=0) = 0.5$. Then we construct linear cases to simplify the analysis that the value of latent dimensions is the combination of factor values.
}

\subsubsection{\revtext{MED vs MIG/SAP.}} 
\revtext{ As mentioned, MIG and SAP require not just \emph{disentanglement} but also \emph{completeness}. So, for a 1000-d model whose latent code value is $c_i = v_{i \text{mod} 2}, 1 \le  i \le D, D = 1000$. I.e. half of the latent code is the true first factor, and the other half latent code is the second factor. It is clear that each dimension perfectly correlates to a single factor. But its MIG/SAP scores turn out to be 0, which is significantly opposite to our intuition. But in this case, for MED we have 
\begin{equation}
    S_i = 1, \quad \rho_i = \frac{1}{D}.
\end{equation}
So $\text{MED}(c)=1$, indicating that every dimension is perfectly disentangled to a single factor and fully responsive to it. Also, this conclusion applies to different values of $D$, showing the robustness of MED with models of different dimensions.
}

\begin{figure}
    \centering
    \includegraphics[width=0.5\textwidth]{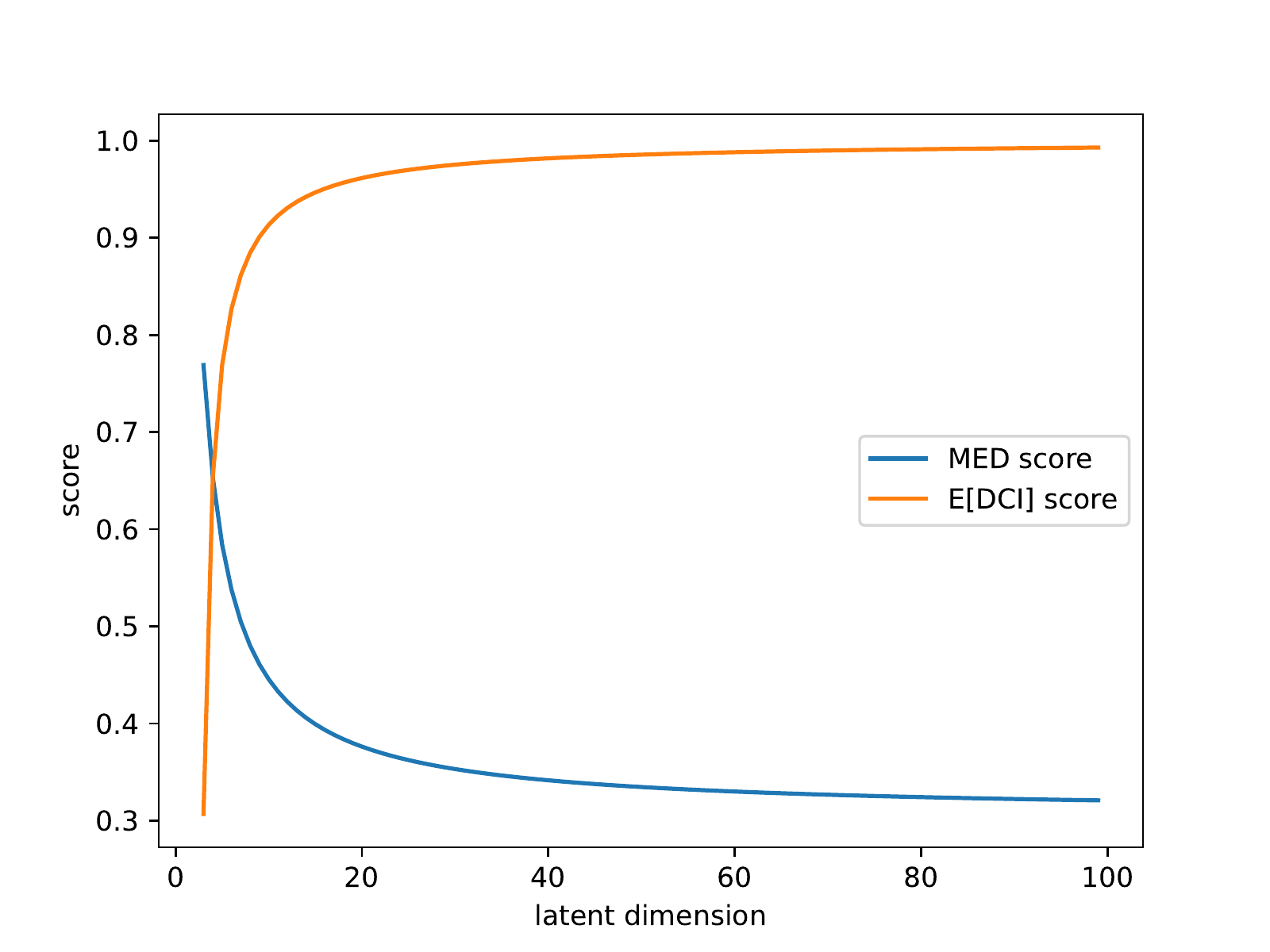}
    \caption{How MED score and DCI score change along the latent dimension of the model in our constructed linear scenarios.}
    \label{fig:med_dci}
\end{figure}

\subsubsection{\revtext{MED vs DCI.}}
\revtext{
With a constructed latent code as 
\begin{equation}
c_i = 
\begin{cases}
  v_0&,i=0\\
  v_1&,i=1\\
  \frac{1}{2}(v_0+v_1) &, i>1
\end{cases}.
\end{equation}
To simplify, we assume the importance matrix output by GBT in DCI metric is the absolute of first-order derivative, namely $|\frac{\partial c_i}{\partial v_j}|$ of the dimension $i$ with respect to the factor $F_j$. GBT encourages sparsity in the output importance matrix. To reflect this, we assume the desired sparsity of the GBT is 2. Thus, we have the dimensions beyond the sparsity limit are given 0 importance. Then the importance matrix value $R_{ij}$ is
\begin{equation}
    R_{ij} = 
    \begin{cases}
    1 &,i=j, i \leq 1 \\
    \frac{1}{2}&, i = d_j\\
    0&,otherwise\\
    \end{cases}.
\end{equation}
$d_j$ is a randomly selected dimension between $[2,D-1]$. For the two factors $F_0$ and $F_1$, $d_j$ is selected independently. So that chance that $d_0 = d_1$ is $p(d_0=d_1) = \frac{1}{D-2}$. 
}

\revtext{
(1) If $d_0=d_1=k$, we have
\begin{equation}
    \rho_i = 
    \begin{cases}
    \frac{1}{3} &, i=0,1,k\\
    0 &, otherwise
    \end{cases}, \quad \quad
    S_i = 
    \begin{cases}
    1 &, i =0,1\\
    1 - log2 &, i=k\\
    0 &, otherwise
    \end{cases}
\end{equation}
}

\revtext{
(2) If $d_0 \neq d_1$, we have
\begin{equation}
    \rho_i = 
    \begin{cases}
    \frac{1}{3} &, i=0,1\\
    \frac{1}{6} &, i=d_0, d_1\\
    0 &, otherwise
    \end{cases}, \quad \quad
    S_i = 
    \begin{cases}
    1 &, i =0,1,d_0,d_1\\
    0 &, otherwise
    \end{cases}.
\end{equation}
And DCI Disentanglement score is $\text{DCI}(c)=\sum_{i=0}^{D-1} \rho_i S_i$. We have $\text{DCI}(c)=1-log2=30.7\%$ in situation (1) and $\text{DCI}(c)=1$ in situation (2). So, in such a simplified version, the expectation of DCI Disentanglement score is 
\begin{equation}
    \mathbb{E}[\text{DCI}(c)] = \frac{1}{D-2} (1-log2)  + (1-\frac{1}{D-2}) = 1 - \frac{log2}{D-2}.
\end{equation}
}

\revtext{On the other hand, similarly we can derive the MED score is
\begin{equation}
    \text{MED}(c)= 1 - \frac{D-2}{D} log2.
\end{equation}
which is quite consistent along the change of $D$. Note that the drastic difference between MED and DCI is caused by the GBT used in DCI. The GBT leads to a sparse importance matrix $M$, while the mutual information in MED won't. This difference doesn't make much difference in low-dimension latent codes, but DCI will be high even if the latent code doesn't disentangle at all in the high-dimensional case. \\
We plot the curves of MED score and the expectation of DCI score in Figure~\ref{fig:med_dci}.
}
\revtext{
It shows that when the latent dimension $D$ increases, the expectation of the DCI score increases and approaches 1. For instance, if $D=3$, we have $\mathbb{E}[\text{DCI}(c)]=30.7\%$; if $D=1000$, we have $\mathbb{E}[\text{DCI}(c)]=99.9\%$. Such an abnormality from DCI Disentanglement alerts us that it can not give a meaningful and fair comparison between methods of different dimensions.
However, MED is much more robust with the change of model dimension. For instance, if $D=3$, we have $\text{MED}(c)=76.9\%$ and if $D=1000$, we have $\text{MED}(c)=30.8\%$.
}

\subsubsection{MED vs BetaVAE/FactorVAE score.}
\textbf{Situation 1} We construct a situation where the first two dimensions are a  weighted combination of the factors and the value of all following dimensions are the average of the factor values. Thus we have the latent code as
\begin{equation}
c_i = 
\begin{cases}
  \frac{1}{3}v_0 + \frac{2}{3}v_1 &,i=0\\
  \frac{1}{3}v_1 + \frac{2}{3}v_0 &,i=1\\
  \frac{1}{2}(v_0+v_1) &, i>1
\end{cases}.
\end{equation}
We can justify it as a very entangled representation. But the latent code value change with respect to the factor value is what BetaVAE and FactorVAE scores care about. 

\revtext{
FactorVAE calculates the variance of latent code, i.e. $Var(c)$, when changing the factors to represent the factor-dimension response degree. During calculating the FactorVAE score, one factor is always fixed. So we have 
\begin{equation}
    Var_{\text{fix}v_0}(c_i) = 
    \begin{cases}
      \frac{1}{9} &, i = 0\\
      \frac{1}{36} &, i=1 \\
      \frac{1}{16} &, i>1\\
    \end{cases}, \quad 
    Var_{\text{fix}v_1}(c_i) = 
    \begin{cases}
      \frac{1}{36} &, i = 0\\
      \frac{1}{9} &, i=1 \\
      \frac{1}{16} &, i>1\\
    \end{cases},
\end{equation}
so for each factor, we have a single dimension that gives the minimum variance. In this case, the FactorVAE score uses a majority vote to determine the dimension with the smallest variance with respect to the value change of a factor as the ``most disentangled'' dimension to it. And the vote accuracy is regarded as the FactorVAE score. So, it is clear that because of the existence of the lucky dimension $j=i$, the FactorVAE score can be expected to approach 100, indicating ``perfectly disentangled''! With the increase of model dimension, the chance of having such a lucky dimension increases as well in real cases. The bias is that FactorVAE only needs one dimension which has a low response to all other factors except for one factor to get high scores.}

\revtext{
On the other hand, the BetaVAE score calculates the latent code value change to represent the response between factors and dimensions. This is represented by the absolute value of the first-order derivative. So for the $i$-th dimension with respect to the $j$-th factor as $R_{ij}$, we have
\begin{equation}
    R_{ij}=
    \begin{cases}
        \frac{1}{3}&,i = j,  i \le 1\\ 
        \frac{2}{3}&,i \neq j, i \le 1 \\
        \frac{1}{2}&,otherwise
    \end{cases}.
\end{equation}}

\revtext{
BetaVAE score uses a learnable linear classifier to determine the dimension. In this case, prediction $j$ from $R_{\cdot j}$ can achieves 100\% accuracy. So similarly, it also requires one ``lucky'' dimension to make the prediction, offering an advantage to high-dimensional models. Moreover, because of the powerful capacity of a learnable linear classifier, the BetaVAE score is close to 100 for many cases as in Table~\ref{table:full_results}. It thus has another flaw of low distinguishability of methods' disentanglement properties. 
}

\revtext{On the other hand, we could now calculate the MED score for this situation. The results turns out to be $\text{MED}(c)=1-log2=30.7\%$ for any value of $D$. This is a relatively low score and we think it makes better sense in such entangled cases compared to the falsely high score from BetaVAE and FactorVAE scores.}

\revtext{\textbf{Situation 2} Let us build another case as in the discussion of DCI above
\begin{equation}
c_i = 
\begin{cases}
  v_0&,i=0\\
  v_1&,i=1\\
  \frac{1}{2}(v_0+v_1) &, i>1
\end{cases}.
\end{equation}
FactorVAE and BetaVAE scores can still give a ``perfect'' score over such a representation. However, when the dimension $D$ is very large, only a small subset of the representation dimensions is disentangled while most are highly entangled. As a higher-dimensional model has a better chance to get a lucky dimension, maintaining the perfect scoring with increasing dimensions is not aligned with our intuition and expectation. From this sense, Top-K MED has its special use. When $D=1000$, we know $\text{MED}(c)=30.8\%$, but we can still get a subset of the 1000 dimensions to get a high Top-K MED score. Such a subset of representations is more compact and has better explainability to the factor variance.}

\revtext{With all the situations constructed and analyzed above, we could notice the failure of existing metrics to achieve evaluation results (1) meaningful, (2) fair, and (3) aligned with the human sense to disentanglement in some scenarios. On the other hand, our proposed MED always keeps the evaluation and comparison reasonable and aligned with the expectation from the human high-level institution.}

\definecolor{gainsboro}{rgb}{0.86, 0.86, 0.86}
\definecolor{azure(web)(azuremist)}{rgb}{0.94, 1.0, 1.0}
\begin{table*}[!t]
\small
\centering
\caption{\revtext{\emph{Logistic Regression} accuracy results of the full representation space and top-$k$ subspace. Methods in \textcolor{gray}{gray} are contrastive self-supervised learning methods.The best results of all models are \textbf{bolded}. The best results on low-dimensional space are \underline{underlined}. }}
\revfig{
\begin{tabular}{c|l|r|r|r|r|r} 
     \toprule
     Space & Model & dSprites & Shapes3D & Cars3D & SmallNORB & CelebA\\ 
     \midrule 
     \multirow{10}{*}{\textbf{Full Space}} & $\beta$-VAE & 34.5 (3.0)& 35.6 (2.6)  & 65.0 (1.6)&  48.6 (0.5) & 82.1 (0.1)\\
     & $\beta$-TCVAE   & 32.1 (3.4) & 40.3 (2.1) & 68.3 (1.4) & 47.4 (0.7) & \utextbf{85.2 (0.1)}\\
     & FactorVAE  & 30.4 (2.5) & 20.7 (2.3) & 56.1 (1.3) & 42.6 (1.3) & 80.4 (0.2)\\
     & DIP-VAE-I  & 30.3 (0.9) & 37.6 (1.5) & \underline{68.5 (1.2)} & \underline{49.6 (0.5)} & \utextbf{85.2 (0.1)}\\
     & DIP-VAE-II& 30.9 (1.9) & -- & 67.2 (1.7) & 48.5 (0.5) & --\\
    \cline{2-7} 
     & \CC{100} MoCo & \CC{100} 46.3 (1.3) & \CC{100} 78.3 (2.1) & \CC{100} 61.8 (1.4) & \CC{100} 56.3 (0.7) & \CC{100} \textbf{85.2 (0.3)} \\
     & \CC{100} MoCov2 & \CC{100} 52.0 (0.3) & \CC{100} 76.5 (0.3) & \CC{100} 63.1 (1.2) & \CC{100} 53.8 (0.5) & \CC{100} 84.9 (0.1) \\
     & \CC{100} BarlowTwins & \CC{100}41.7 (0.2) & \CC{100} 74.5 (0.8)  & \CC{100} 79.0 (0.2) & \CC{100}\textbf{62.0 (0.1)} & \CC{100} \CC{100} 83.1 (0.2) \\ 
     & \CC{100} SimSiam & \CC{100} 40.3 (0.1) & \CC{100} \textbf{80.9 (3.4)} & \CC{100} 45.3 (4.6) & \CC{100} 54.5 (0.5)  & \CC{100} 84.0 (0.2)\\
     & \CC{100} BYOL  & \CC{100} \textbf{54.6 (0.2) } & \CC{100} 71.5 (1.3)& \CC{100} \textbf{81.2 (0.4)} & \CC{100} 57.2 (1.1) & \CC{100} 81.9 (0.2)\\
     \cmidrule(r){1-7}
     \multirow{5}{*}{\makecell{\textbf{Top-$k$} \textbf{Subspace}}}   &  \CC{100} MoCo &  \CC{100} 22.7 (1.4) & \CC{100}43.7 (1.1) & \CC{100} 38.1 (1.2) & \CC{100} 34.5 (0.2) & \CC{100} 85.1 (0.2) \\
     &  \CC{100} MoCov2 & \CC{100} 26.0 (1.0) & \CC{100}33.4 (0.2) & \CC{100} 32.6 (0.8) & \CC{100} 31.4 (1.1) & \CC{100} 84.9 (0.1) \\
     &  \CC{100} BarlowTwins & \CC{100} 29.1 (0.7) & \CC{100}44.4 (1.3)  & \CC{100} 43.1 (1.4) & \CC{100} 34.5 (0.4) &\CC{100} 82.9 (0.2) \\
     &  \CC{100} SimSiam &  \CC{100} \underline{38.0 (0.6)} & \CC{100} \underline{61.8 (1.0)} & \CC{100} 29.4 (3.6) & \CC{100} 37.7 (2.5) & \CC{100} 83.8 (0.2)\\
     &  \CC{100}  BYOL  & \CC{100} 37.9 (0.6) & \CC{100} 46.3 (0.5) & \CC{100} 43.3 (0.9) & \CC{100} 40.2 (0.8) & \CC{100} 81.9 (0.2)\\
     \bottomrule
\end{tabular}
}
\label{table:downstreams}
\end{table*}

\section{Properties of the Top-k Subspace}
We propose the top-$k$ selection process to study the partial disentangle properties. Here we provide more details. We first visualize the top-$k$ selection process based on the heatmap of the importance matrix. Then we will discuss the effects of $k$ and the downstream performance comparison of the top-$k$ subspace and full space.

\begin{figure}[h]
    \centering 
    \revfig{\includegraphics[width=\linewidth]{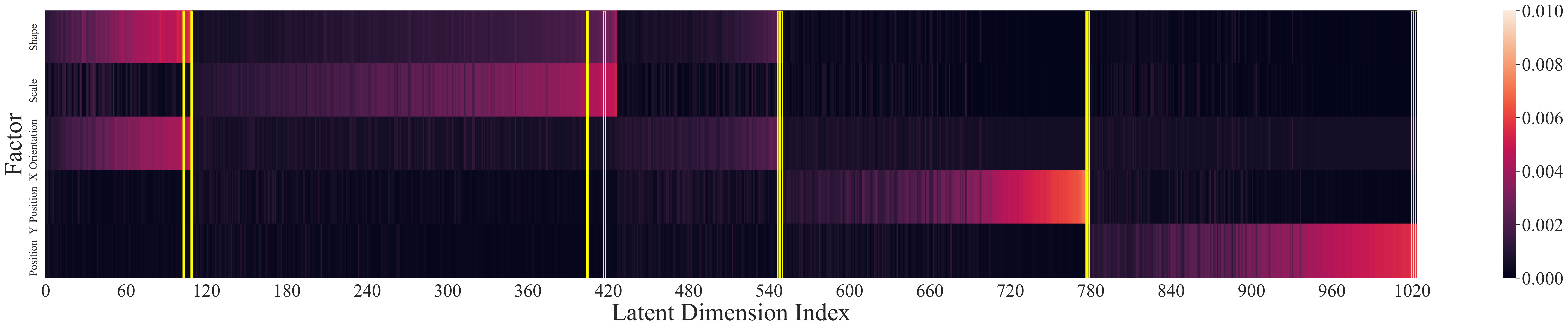}}
    \caption{\revtext{The visualization of the top-$k$ process. Based on Figure \ref{fig:mi_dsprites_full}, the selected dimensions are highlighted with \textcolor{yellow}{yellow} boxes. For dSprites we set $k=2$. The indices picked are 103, 109, 404, 417, 547, 549, 777, 778, 1020, and 1023.}}
    \label{fig:mi_dsprites_with_box}
\end{figure}

\paragraph{Visualization of the top-$k$ Process} We describe the top-$k$ process in the Section \ref{sec:partial_med}. Correspondingly, here we colored the picked dimensions \textcolor{yellow}{yellow} in the Figure \ref{fig:mi_dsprites_with_box}.
It shows that most picked dimensions only have one entry much brighter than the others, indicating they are well-disentangled. Since \emph{orientation} is ill-defined in dSprites, BYOL tends to encode it with \emph{shape}. Hence the most disentangled dimensions for \emph{shape} also capture information of \emph{orientation}.  

\begin{figure}[h]
    \centering
    \includegraphics[width=0.5\textwidth]{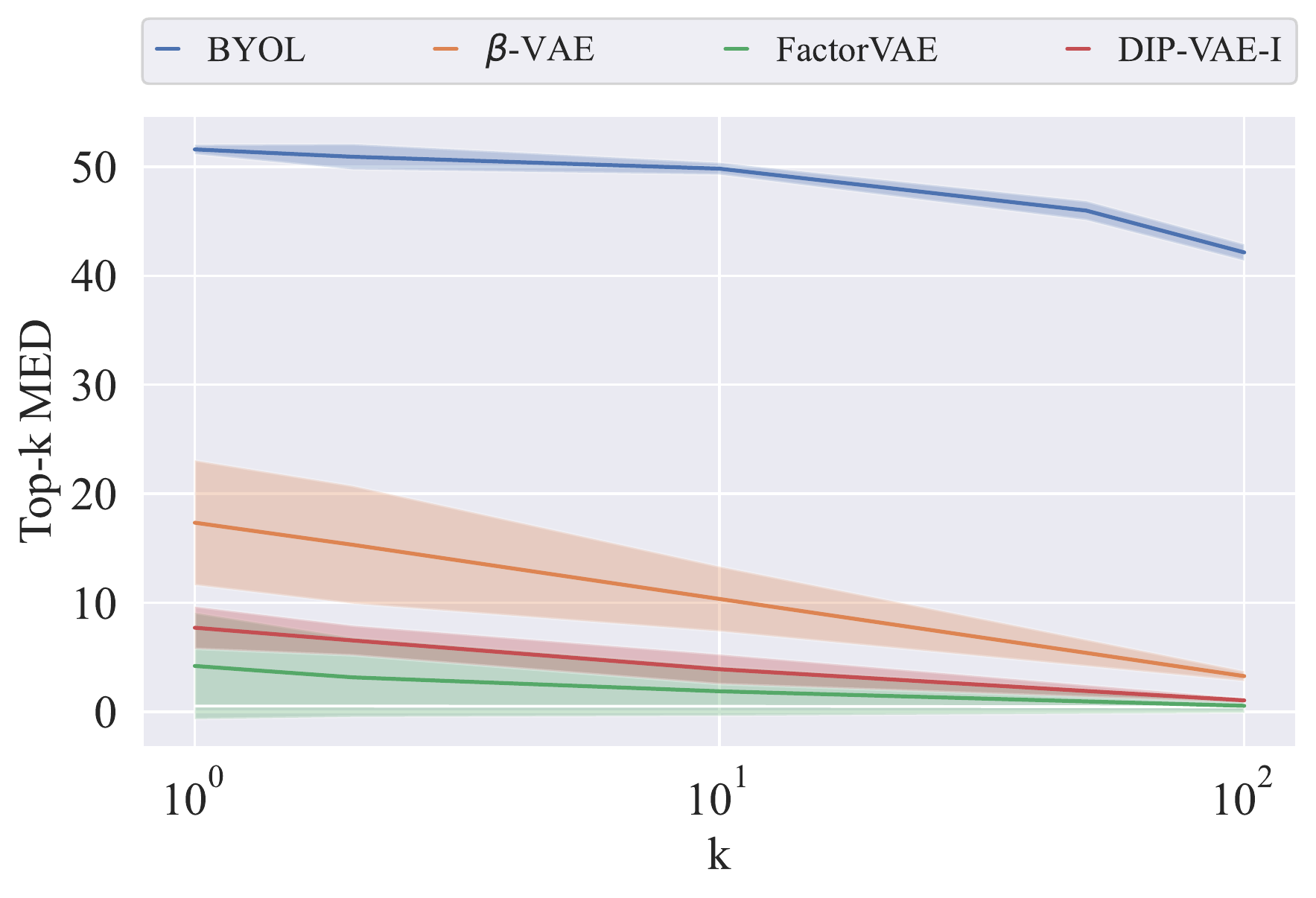}
    \caption{\revtext{The influence of $k$ in top-$k$ MED on dSprites dataset. The evaluation results are consistent with multiple choices of $k$.} }
    \label{fig:k_ablation}
\end{figure}

\paragraph{Ablation on $k$}  The parameter $k$ is designed as an evaluation choice. For analytical purposes, we choose the value of $k$ to match the dimension of VAE methods in the main text. To investigate the influence of $k$, we set $k=1, 2, 10, 50, 100$ and evaluate the top-$k$ MED scores of 1000-d VAEs and BYOL on dSprites. The results are shown in Figure \ref{fig:k_ablation}. We find that with different values of $k$, the ranks of top-$k$ MED scores remain unchanged. Since the top-$k$ process is a supervised greedy selection, as $k$ grows the scores decrease. In conclusion, top-$k$ MED can measure the subspace disentangle properties with a wide range of $k$. Therefore we can choose $k$ according to actual needs.

\paragraph{Downstream performance of top-$k$ subspace} We study the disentangle properties of the full space and top-$k$ subspace in \ref{sec:benchmark}. Here we analyze the downstream performance of these spaces. We train a \emph{Logistic Regressor} to predict each factor from the full representation or the top-$k$ picked representation. Then we take the mean prediction accuracy of all factors as the measure of downstream performance. We compare the downstream performance of the full space of VAEs and BYOL,  and top-$k$ subspace of BYOL on multiple datasets. Here we set $k=2$ for dSprites, Shape3D, and SmallNORB, and $k = 3$ for Cars3D and CelebA to match the dimension of VAEs.  The results are in Table \ref{table:downstreams}. We note that, except on CelebA, performance drops non-negligibly when completing downstream tasks with the top-$k$ representations. This suggests that the top-$k$ subspace is well-disentangled but other dimensions also contain factor-related information. It is unsurprising because we cut down the dimension drastically (from 1000 to 10 on synthetic datasets and from 1000 to 100 on CelebA). However, we find that the top-$k$ spaces learned by contrastive methods achieve better or compatible performance compared with the full spaces of VAEs on 3 out of 5 datasets (dSprites, Shapes3D, and CelebA). It is reasonable that a high-dimensional model can be better for downstream tasks as the downstream model can be trained to leverage each little piece of information from the input representations. But we get a message from the results that by drawing a subspace from the powerful CL-trained model, its performance on downstream tasks can be comparable to or superior to the low-dimensional models. And the selected low-dimensional model can have much better compactness and interpretability.

\section{Limitations}
Our work still has some limitations. Firstly, the design of contrastive learning methods still depends on empirical practice. Therefore there lacks a widely accepted theoretical framework to analyze its disentanglement properties. As for the experiment setups, the inductive biases of different methods may potentially affect the results (such as normalization in Section~\ref{sec:appendix_ablation}). Instead of an exhaustive search, we inherit the available best settings, i.e., the settings from DisLib~\cite{locatello2019challenging} and the public official implementations of other methods. All hyperparameters and details have been indicated in Section~\ref{sec:reproducibility}.

\end{document}